\documentclass[A4,11pt]{article}

\usepackage[dvips]{color}

\usepackage{times}
\usepackage{latexsym}

\usepackage{psfig}
\usepackage{graphicx}
\usepackage{pstricks}
\usepackage{amssymb}
\usepackage{fullname}

\def\reportnumber{2004$\,${\it /}$\,$031}
\def\reportdate{June 2, 2004}

\input{ntape.def}

\begin{document}


\title{
  Algorithms for Weighted Multi-Tape Automata \\
  {\Large --~ XRCE Research Report \reportnumber ~~--}
  \vspace{2ex}
}

\author{
  Andre Kempe$^{1}$\spc{4ex}
  Franck Guingne$^{1,2}$\spc{4ex}
  Florent Nicart$^{1,2}$
  \vspace{2ex} \\
  $^{1}$
  Xerox Research Centre Europe -- Grenoble Laboratory \\
  6 chemin de Maupertuis -- 38240 Meylan -- France \\
  {\small\sf firstname.lastname@xrce.xerox.com}
	~--~ {\small\sf http://www.xrce.xerox.com}
  \vspace{2ex} \\
  $^{2}$
  Laboratoire d'Informatique Fondamentale et Appliqu\'ee de Rouen \\
  Facult\'e des Sciences et des Techniques -- Universit\'e de Rouen \\
  76821 Mont-Saint-Aignan -- France \\
  {\small\sf firstname.lastname@dir.univ-rouen.fr}
	~--~ {\small\sf http://www.univ-rouen.fr/LIFAR/}
  \vspace{3ex}
}

\date{\reportdate}

\maketitle

\thispagestyle{empty}


\vspace{5ex}

\section*{Abstract}

This report defines various operations and describes algorithms
for {\it weighted multi-tape automata}\/  \linebreak
(WMTAs).
It presents, among others, a new approach to {\it multi-tape intersection}\/,
meaning the intersection of a number of tapes of one WMTA
with the same number of tapes of another WMTA,
which can be seen as a generalization of transducer intersection.
In our approach,
multi-tape intersection is not considered as an atomic operation
but rather as a sequence of more elementary ones.
We show an example of multi-tape intersection,
actually transducer intersection,
that can be compiled with our approach
but not with several other methods that we analyzed.
Finally we describe an example of practical application,
namely the preservation of intermediate results in transduction cascades.


\newpage

\tableofcontents


\newpage

\renewcommand{\RSec}[1]{\section{#1}}
\renewcommand{\RSubSec}[1]{\subsection{#1}}
\renewcommand{\RSubSubSec}[1]{\subsubsection{#1}}

\RSec{Introduction
  \label{sec:intro}}

Finite state automata (FSAs) and
weighted finite state automata (WFSAs)
are well known, mathematically well defined, and offer many practical advantages.
\cite{elgot+mezei:1965,eilenberg:1974,kuich+salomaa:1986}.
They permit, among others, the fast processing of input strings
and can be easily modified and combined by well defined operations.
Both FSAs and WFSAs are widely used in language and speech processing
\cite{kaplan+kay:1981,koskenniemi:1992,sproat:1992,karttunen+al:1997,mohri:1997,roche+schabes:1997}.
A number of software systems have been designed to manipulate FSAs and WFSAs
\cite{karttunen+al:1997,vannoord:1997,mohri+al:1998,beesley+karttunen:2003}.
Most systems and applications deal, however, only with {\it 1-tape}\/ and
{\it 2-tape automata}\/,
also called acceptors and transducers, respectively.

{\it Multi-tape automata}\/ (MTAs)
\cite{elgot+mezei:1965,kaplan+kay:1994}
offer additional advantages such as the possibility of
storing different types of information, used in NLP, on different tapes
or preserving intermediate results of transduction cascades on different tapes
so that they can be re-accessed by any of the following transductions.
MTAs have been implemented and used, for example,
in the morphological analysis of Semitic languages,
where the vowels, consonants, pattern, and surface form
of words have been represented on different tapes of an MTA
\cite{kay:1987,kiraz:1997,kiraz+grimley-evans:1998}.

This report defines various operations for {\it weighted multi-tape automata}\/ (WMTAs)
and describes algorithms that have been implemented for those operations
in the WFSC toolkit
\cite{kempe+al:2003}.
Some algorithms are new, others are known or similar to known algorithms.
The latter will be recalled to make this report more complete and self-standing.
We present a new approach to {\it multi-tape intersection}\/,
meaning the intersection of a number of tapes of one WMTA
with the same number of tapes of another WMTA.
In our approach,
multi-tape intersection is not considered as an atomic operation
but rather as a sequence of more elementary ones,
which facilitates its implementation.
We show an example of multi-tape intersection,
actually transducer intersection,
that can be compiled with our approach
but not with several other methods that we analyzed.
To show the practical relevance of our work,
we include an example of application:
the preservation of intermediate results in transduction cascades.

For the structure of this report see the table of contents.


\renewcommand{\RSec}[1]{\section{#1}}
\renewcommand{\RSubSec}[1]{\subsection{#1}}
\renewcommand{\RSubSubSec}[1]{\subsubsection{#1}}

\RSec{Some Previous Work
	\label{sec:Previous}}

\RSubSec{$n$-Tape Automaton Seen as a Two-Tape Automaton}

\namecite{rabin+scott:1959}
presented in a survey paper a number of results and problems on finite 1-way automata,
the last of which
-- the decidability of the equivalence of deterministic k-tape automata --
has been solved only recently and by means of purely algebraic methods
\cite{harju+karhumaki:1991}.

Rabin and Scott
considered the case of two-tape automata claiming this is not a loss of generality. 
They adopted the convention
``$\dots$ that the machine will read  for a while on one tape, 
then change control and read a while on the other tape, 
and so on until one of the tapes is exhausted $\dots$".
In this view, a two-tape or $n$-tape machine is just an ordinary automaton 
with a partition of its states to determine which tape is to be read.

\pagebreak  

\RSubSec{$n$-Tape Automaton Seen as a Single-Tape Automaton}

\namecite{ganchev+al:2003}
define the notion of ``one-letter $k$-tape automaton''
and the main idea is to consider this restricted form of $k$-tape automata 
where all transition labels have exactly one tape with a non-empty single letter. 
Then they prove that one can use ``classical'' algorithms for 1-tape automata 
on a one-letter $k$-tape automaton. 
They propose an additional condition to be able to use classical intersection. 
It is based on the notion that a tape or coordinate is {\it inessential}\/ 
iff $\forall \aTuple{ w_1, ...,w_k} \in R$ ~($R$ is a regular relation over $(\Sigma^*)^k$)
and $\forall v \in \Sigma^*$, $\aTuple{ w_1, ...w_{i-1}, v, w_{i+1}, ...,w_k} \in R$. 
And thus to perform an intersection,
they assume that there exists at most one common essential tape between the two operands.

\RSubSec{$n$-Tape Transducer}

\namecite{kaplan+kay:1994}
define a non-deterministic {\it $n$-way finite-state transducer}\/ 
that is similar to a classic transducer except that the transition function maps
$Q \times \Sigma^\epsilon \times ... \times \Sigma^\epsilon$ to $2^Q$ 
~(with $\Sigma^\eps = \Sigma \cup \{\eps\}$). 
To perform the {\it intersection}\/  between two $n$-tape transducers, 
they introduced the notion of {\it same-length relations}\/ . 
As a result, they treat a subclass of $n$-tape transducers to be intersected.

\namecite{kiraz:1997}
defines an $n$-tape finite state automaton and an {\it $n$-tape finite-state transducer}\/,
introducing the notion of {\it domain  tape}\/ and {\it range tape}\/
to be able to define a unambiguous composition for $n$-tape transducers. 
Operations on $n$-tape automata are based on \cite{kaplan+kay:1994}\/ , 
the intersection in particular.


\renewcommand{\RSec}[1]{\section{#1}}
\renewcommand{\RSubSec}[1]{\subsection{#1}}
\renewcommand{\RSubSubSec}[1]{\subsubsection{#1}}

\RSec{Mathematical Objects
	\label{sec:Obj}}

In this section we recall the basic definitions of the algebraic structures
monoid and semiring,
and give a detailed definition of a weighted multi-tape automaton (WMTA)
based on the definitions of a weighted automaton and a multi-tape automaton
\cite{rabin+scott:1959,elgot+mezei:1965,eilenberg:1974,kuich+salomaa:1986}.


\RSubSec{Semirings
		\label{sec:Obj:Sem}}

A {\it monoid}\/ is a structure $\aTuple{M,\anyOp,\srOne}$
consisting of a set $M$, an associative binary operation $\anyOp$ on $M$,
and a {\it neutral}\/ element $\srOne$ such that
$ \srOne \anyOp a = a \anyOp \srOne = a $ for all $a\!\in\! M$.
A monoid is called {\it commutative}\/ iff
$ a \anyOp b = b \anyOp a $ for all $a,b\!\in\! M\;$.

A set $\srSetK$ equipped with two binary operations,
$\srPlus$ ({\it collection}\/) and $\srTimes$ ({\it extension}\/),
and two neutral elements, $\srZero$ and $\srOne$, is called a {\it semiring}\/,
iff it satisfies the following properties:
\begin{enumerate} 
\item $\aTuple{\srSetK,\srPlus, \srZero}$ is a commutative monoid
\item $\aTuple{\srSetK,\srTimes, \srOne}$ is a monoid
\item 
extension is {\it left-}\/ and {\it right-distributive}\/ over collection: \\
$a\srTimes (b\srPlus c) = (a\srTimes b) \srPlus (a\srTimes c)\,,\;\;
 (a\srPlus b)\srTimes c = (a\srTimes c) \srPlus (b \srTimes c)\,,\;\;
\forall a,b,c\!\in\! \srSetK$
\item $\srZero$ is an annihilator for extension:~~
$\srZero\srTimes a=a\srTimes \srZero=\srZero \,,\;\; \forall a\!\in\! \srSetK$
\end{enumerate}

\noindent
We denote a generic semiring as
$\srK = \aTuple{\srSetK, \srPlus, \srTimes, \srZero, \srOne}$.

Some automaton algorithms require semirings to have specific properties. 
Composition, for example, requires it to be commutative 
\cite{pereira+al:1997,mohri+al:1998} 
and $\eps$-removal requires it to be {\it k-closed}\/
\cite{mohri:2002}. 
These properties are defined as follows: 
\begin{enumerate} 
\item commutativity:~ 
$ a \srTimes b = b \srTimes a \;,\;\; \forall a,b\!\in\! \srSetK $ 
\item k-closedness:~ 
$ \srBigPlus_{n=0}^{k+1} a^n = \srBigPlus_{n=0}^{k} a^n \;,\;\;
	\forall a\!\in\! \srSetK $ 
\end{enumerate} 

\noindent
The following well-known semirings are commutative:
\begin{enumerate}
\item
$\srB = \aTuple{\srSetB, \logOr, \logAnd, 0, 1}$~:
	the boolean semiring, with $\srSetB=\aSet{0,1}$
\item
$\srN = \aTuple{\srSetN, +, \times , 0, 1}$~:
	a positive integer semiring with arithmetic addition and multiplication
\item
$\srR^+ = \aTuple{\srSetR^+, +, \times, 0, 1 }$~:
	a positive real semiring
\item
$\overline{\srR}^+ = \aTuple{\overline{\srSetR}^{\;+} , min, +, \infty, 0 }$~:
	a real tropical semiring, with $ \overline{\srSetR}^{\;+} = \srSetR^+ \cup \{\infty \}$
\end{enumerate}

A number of algorithms require semirings to be equipped
with an order or partial order denoted by $<_{\srK}$.
Each idempotent semiring $\srK$
(i.e., $\forall a\!\in\!\srK: a \srPlus a = a$)
has a natural partial order defined by
$a <_{\srK} b \Leftrightarrow a \srPlus b = a$.
In the above examples, the boolean and the real tropical semiring
are idempotent, and hence have a natural partial order.


\RSubSec{Weighted Multi-Tape Automata
		\label{sec:Obj:Wmta}}

In analogy to a weighted automaton and a multi-tape automaton (MTA),
we define a {\it weighted multi-tape automaton}\/ (WMTA),
also called weighted $n$-tape automaton,
over a semiring $\srK$, as a six-tuple
\begin{equation}
A\tapnum{n}\DefAs\aTuple{\Sigma, Q, I, F, E\tapnum{n}, \srK}
\end{equation}

\noindent
with

\spc{8mm}\begin{definitionsTTT}{20mm}{42mm}{90mm}
$\Sigma$	&	& being a finite alphabet		\\
$Q$	&		& the finite set of states		\\
$I$	& $\subseteq Q$	& the set of initial states		\\
$F$	& $\subseteq Q$	& the set of final states		\\
$n$	&		& the arity, i.e., the number of tapes of $A\tapnum{n}$	\\
$E\tapnum{n}$	& $\subseteq Q \times (\Sigma^*)^n \times \srSetK \times Q$	&
			  being the finite set of $n$-tape transitions and	\\
$\srK$	& $=\aTuple{\srSetK,\srPlus,\srTimes,\srZero,\srOne}$	&
			  the semiring of weights.		\\
\end{definitionsTTT}

\noindent
For any state $q\in Q$,

\spc{8mm}\begin{definitionsTTT}{20mm}{42mm}{90mm}
$\wgtInit(q)$	& $\in\srK$	&
	denotes its initial weight, with $\wgtInit(q) \not= \srZero \Leftrightarrow q \in I$,	\\
$\wgtFin(q)$	& $\in\srK$ 	&
	its final weight, with $\wgtFin(q) \not= \srZero \Leftrightarrow q \in F$, and	\\
$E(q)$		& $\subseteq E\tapnum{n}$	&
	its finite set of out-going transitions.	\\
\end{definitionsTTT}

\noindent
For any transition $e\tapnum{n} \in E\tapnum{n}$, with $e\tapnum{n}\!=\!\aTuple{p,\lab\tapnum{n},w,n}$,

\spc{8mm}\begin{definitionsTTT}{20mm}{42mm}{90mm}
$p(e\tapnum{n})$ 	& $p : E\tapnum{n} \rightarrow Q$	&
	denotes its source state	\\
$\lab(e\tapnum{n})$	& $\lab : E\tapnum{n} \rightarrow (\Sigma^*)^n$	&
	its label, which is an $n$-tuple of strings	\\
$w(e\tapnum{n})$	& $w : E \rightarrow \srK$	&
	its weight, with $w(e\tapnum{n}) \not= \srZero \Leftrightarrow e\tapnum{n} \in E\tapnum{n}$, and	\\
$n(e\tapnum{n})$	& $n : E \rightarrow Q$	&
	its target state	\\
\end{definitionsTTT}

A {\it path} $\path\tapnum{n}$ of length $r\!=\!|\path\tapnum{n}|$
is a sequence of transitions $e_1\tapnum{n} e_2\tapnum{n} \cdots e_r\tapnum{n}$
such that $n(e_i\tapnum{n})\!=\!p(e_{i+1}\tapnum{n})$
for all $i\!\in\!\rangeL 1, r\!-\!1 \rangeR$.
A path is said to be {\it successful}
iff $p(e_1\tapnum{n})\!\in\! I$ and $n(e_r\tapnum{n})\!\in\! F$.
In the following we consider only successful paths.
The label of a successful path $\path\tapnum{n}$
equals the concatenation of the labels of its transitions
\begin{equation}
\plab(\path\tapnum{n})	\;\;=\;\;
  \lab(e_1\tapnum{n}) \; \lab(e_2\tapnum{n}) \; \cdots \; \lab(e_r\tapnum{n})
\end{equation}

\noindent
and is an $n$-tuple of strings
\begin{equation}
\plab(\path\tapnum{n}) \;\;=\;\; s\tapnum{n} \;\;=\;\; \aTuple{s_1, s_2, \dots , s_n}
\end{equation}

\noindent
If all strings $s_j \in \Sigma^*$ (with $j \in \rangeL 1 , n \rangeR$)
of a tuple $s\tapnum{n}$ are equal,
we use the short-hand notation $s_j\tapnum{n}$ on the terminal string $s_j$.
For example:
\begin{eqnarray}
  (abc)\tapnum{3} & = & \aTuple{abc, abc, abc}  \\
  \eps\tapnum{4}  & = & \aTuple{\eps, \eps, \eps, \eps}
\end{eqnarray}

\noindent
The $n$ strings on any transition $e\tapnum{n}$ are not ``bound'' to each other.
For example, the string triple $s\tapnum{3}\!=\!\aTuple{aaa, bb, cccc}$
can be encoded, among others, by any of the following sequences of transitions:
$(a{\T}b{\T}cc)(a{\T}b{\T}c)(a{\T}\eps{\T}c)$ or
$(aa{\T}\eps{\T}\eps)(a{\T}b{\T}cc)(\eps{\T}b{\T}cc)$ or
$(aaa{\T}bb{\T}cccc)(\eps{\T}\eps{\T}\eps)$, etc.

The weight $w(\path\tapnum{n})$ of a successful path is
\begin{equation}
w(\; \path\tapnum{n} \;)	\;\;=\;\;
	\wgtInit(\; p(e_1\tapnum{n}) \;) \;\srTimes\;
	\left( \srBigTimes_{j=\rangeL 1,r \rangeR}w(\;e_j\tapnum{n}\;) \right) \;\srTimes\;
	\wgtFin(\; n(e_r\tapnum{n}) \;)
\end{equation}

We denote by $\Paths(A\tapnum{n})$
the (possibly infinite) set of successful paths of $A\tapnum{n}$
and by $\Paths(s\tapnum{n})$ the (possibly infinite) set of successful paths
for the $n$-tuple of strings $s\tapnum{n}$
\begin{equation}
  \Paths(s\tapnum{n})	\;=\;
	\aSet{\; \path\tapnum{n}\!\in\!\Paths(A\tapnum{n}) \;|\;
			s\tapnum{n}\!=\!\plab(\path\tapnum{n}) \;}
\end{equation}

We call $\relat(A\tapnum{n})$ the $n$-ary or $n$-tape relation of $A\tapnum{n}$.
It is the (possibly infinite) set of $n$-tuples of strings $s\tapnum{n}$
having successful paths in $A\tapnum{n}$:
\begin{equation}
  \relat\tapnum{n} \;\;=\;\;
  \relat(A\tapnum{n}) \;\;=\;\;
	\aSet{\; s\tapnum{n} \;|\;
		\exists \path\tapnum{n}\!\in\!\Paths(A\tapnum{n}) \wedge \plab(\path\tapnum{n})=s\tapnum{n} \;}
\end{equation}

\noindent
The weight for any $n$-tuple of strings $s\tapnum{n}\!\in\!\relat(A\tapnum{n})$
is the collection (semiring sum) of the weights of all paths labeled with $s\tapnum{n}$~:
\begin{equation}
w(s\tapnum{n}) \;\;=\;\;
	\srBigPlus_{\path\tapnum{n}\in \Paths\left(s\tapnum{n}\right)} w(\path\tapnum{n})
\end{equation}

By relation we mean simply a co-occurrence of strings in tuples.
We do not assume any particular relation between those strings
such as an input-output relation.
All following operations and algorithms are independent
from any particular relation.
It is, however, possible to define an arbitrary weighted relation
between the different tapes of $\relat(A\tapnum{n})$.
For example,
$\relat(A\tapnum{2})$ of a weighted {\it transducer}\/ $A\tapnum{2}$
is usually considered as a weighted input-output relation between its two tapes,
that are called {\it input tape}\/ and {\it output tape}\/.

In the following we will not distinguish between a language ${\cal L}$
and a 1-tape relation $\relat\tapnum{1}$,
which allows us to define operations only on relations
rather than on both languages and relations.


\renewcommand{\RSec}[1]{\section{#1}}
\renewcommand{\RSubSec}[1]{\subsection{#1}}
\renewcommand{\RSubSubSec}[1]{\subsubsection{#1}}

\RSec{Operations
	\label{sec:Op}}

This section defines operations on string $n$-tuples and $n$-tape relations,
taking their weights into account.
Whenever these operations are used on transitions, paths, or automata,
they are actually applied to their labels or relations respectively.
For example, the binary operation $\anyBinOp$ on two automata,
$A_1\tapnum{n} \anyBinOp A_2\tapnum{n}$,
actually means
$\relat(A_1\tapnum{n}\anyBinOp A_2\tapnum{n})=\relat(A_1\tapnum{n})\anyBinOp\relat(A_2\tapnum{n})$.
The unary operation $\anyUniOp$ on one automaton, $\anyUniOp A\tapnum{n}$,
actually means
$\relat(\anyUniOp A\tapnum{n})=\anyUniOp\relat(A\tapnum{n})$.

Ultimately, we are interested in multi-tape intersection and transduction.
The other operations are introduced because they serve as basis for the two.


\RSubSec{Pairing and Concatenation}

We define the {\it pairing} of two string tuples,
$s\tapnum{n} \paired v\tapnum{m} = u\tapnum{n+m}$,
and its weight as
\begin{eqnarray}
  \aTuple{s_1, \dots , s_n} \paired \aTuple{v_1, \dots , v_m}
	& \DefAs & \aTuple{s_1, \dots , s_n, v_1, \dots , v_m}  \\
  w\left(\; \aTuple{s_1, \dots , s_n} \paired \aTuple{v_1, \dots , v_m} \;\right)
	& \DefAs & w\left(\; \aTuple{s_1, \dots , s_n} \;\right)
				\srTimes w\left(\; \aTuple{v_1, \dots , v_m} \;\right)
	\label{eq:Op:PairingWeight}
\end{eqnarray}

\noindent
Pairing is associative (concerning both the string tuples and their weights)~:
\begin{equation}
  s_1\tapnum{n_1}{\T}s_2\tapnum{n_2}{\T}s_3\tapnum{n_3} \;=\;
  \left(s_1\tapnum{n_1}{\T}s_2\tapnum{n_2}\right){\T}s_3\tapnum{n_3} \;=\;
  s_1\tapnum{n_1}{\T}\left(s_2\tapnum{n_2}{\T}s_3\tapnum{n_3}\right) \;=\;
  s\tapnum{n_1+n_2+n_3}
\end{equation}

\noindent
We will not distinguish between 1-tuples of strings and strings, and hence,
instead of $s\tapnum{1}{\T}v\tapnum{1}$ or $\aTuple{s}{\T}\aTuple{v}$,
simply write $s{\T}v$.

The {\it concatenation} of two string tuples of equal arity,
$s\tapnum{n} v\tapnum{n} = u\tapnum{n}$,
and its weight are defined as
\begin{eqnarray}
  \aTuple{s_1, \dots , s_n} \aTuple{v_1, \dots , v_n}
	& \DefAs & \aTuple{s_1 v_1, \dots , s_n v_n}  \\
  w\left(\; \aTuple{s_1, \dots , s_n} \aTuple{v_1, \dots , v_n} \;\right)
	& \DefAs & w\left(\; \aTuple{s_1, \dots , s_n} \;\right)
				\srTimes w\left(\; \aTuple{v_1, \dots , v_n} \;\right)
\end{eqnarray}

\noindent
Concatenation is associative (concerning both the string tuples and their weights)~:
\begin{equation}
  s_1\tapnum{n} s_2\tapnum{n} s_3\tapnum{n} \;=\;
  \left(s_1\tapnum{n} s_2\tapnum{n}\right) s_3\tapnum{n} \;=\;
  s_1\tapnum{n} \left(s_2\tapnum{n} s_3\tapnum{n}\right) \;=\;
  s\tapnum{n}
\end{equation}

\noindent
Again, we will not distinguish between 1-tuples of strings and strings, and hence,
instead of $s\tapnum{1} v\tapnum{1}$ or $\aTuple{s} \aTuple{v}$,
simply write $s v$.

The relation retween pairing and concatenation can be expressed
through a matrix of string tuples
\begin{equation}
  \left[
  \begin{array}{ccc}
    s_{11}\tapnum{n_1}	&  \cdots	&  s_{1r}\tapnum{n_1}	\\
		\vdots		&		&	\vdots			\\
    s_{m1}\tapnum{n_m}	&  \cdots	&  s_{mr}\tapnum{n_m}	\\
  \end{array}
  \right]
\end{equation}

\noindent
where the $s_{jk}\tapnum{n_j}$ are horizontally concatenated and vertically paired:
\begin{eqnarray}
  s\tapnum{n_1+\dots+n_m}
  & = &
  \left( s_{11}\tapnum{n_1} \;\cdots\; s_{1r}\tapnum{n_1} \right)
  \;\;\paired\;\; \cdots \;\;\paired\;\;
  \left( s_{m1}\tapnum{n_m} \;\cdots\; s_{mr}\tapnum{n_m} \right)  \nonumber \\
  & = &
  \left( s_{11}\tapnum{n_1} \;\;\paired\;\; \cdots \;\;\paired\;\; s_{m1}\tapnum{n_m} \right)
  \cdots
  \left( s_{1r}\tapnum{n_1} \;\;\paired\;\; \cdots \;\;\paired\;\; s_{mr}\tapnum{n_m} \right)
\end{eqnarray}

\noindent
Note, this equation does not hold for the weights of the $s_{jk}\tapnum{n_j}$,
unless they are defined over a commutative semiring $\srK$.


\RSubSec{Cross-Product
		\label{sec:Op:CrProd}}

The {\it cross-product}\/ of two $n$-tape relations,
$\relat_1\tapnum{n}\!\CrPr\!\relat_2\tapnum{m}\!=\!\relat\tapnum{n+m}$,
is based on pairing and is defined as
\begin{equation}
  \relat_1\tapnum{n} \CrPr \relat_2\tapnum{m}
	\; \DefAs \; \aSet{\; s\tapnum{n} \paired v\tapnum{m} \;\;|\;\;
		s\tapnum{n}\in\relat_1\tapnum{n},
		v\tapnum{m}\in\relat_2\tapnum{m} \;}
	\label{eq:Op:CrProd}
\end{equation}

\noindent
The weight of each string tuple
$u\tapnum{n+m}\in\relat_1\tapnum{n}\CrPr\relat_2\tapnum{m}$
follows from the definition of pairing.

The cross product is an associative operation.

A well-know special case is the cross-product of two acceptors (1-tape automata)
leading to a transducer (2-tape automaton)~:
\begin{eqnarray}
  A\tapnum{2}
	& = & A_1\tapnum{1} \CrPr A_2\tapnum{1}  \\
  \relat(\; A\tapnum{2} \;)
	& = & \aSet{\; s \paired v \;\;|\;\;
		s\in\relat(A_1\tapnum{1}),
		v\in\relat(A_2\tapnum{1}) \;}  \\
  w_A(\; s \paired v \;)
	& = & w_{A_1}(s) \;\srTimes\; w_{A_2}(v)
\end{eqnarray}


\RSubSec{Projection and Complementary Projection
  \label{sec:Op:Projection}}

The {\it projection}\/, $\Proj{j,k,\dots}(s\tapnum{n})$,
of a string tuple is defined as
\begin{equation}
  \Proj{j,k,\dots}(\; \aTuple{s_1, \dots , s_n} \;) \DefAs \aTuple{s_j, s_k, \dots}
\end{equation}
\noindent
It retains only those strings (i.e., tapes) of the tuple
that are specified by the indices $j,k,\dots \in\rangeL 1, n \rangeR$,
and places them in the specified order.
Projection indices can occur in any order and more that once.
Thus the tapes of $s\tapnum{n}$ can, e.g., be reversed or duplicated:
\begin{eqnarray}
  \Proj{n,\dots,1}(\; \aTuple{s_1, \dots , s_n} \;) & = & \aTuple{s_n, \dots , s_1}  \\
  \Proj{j,j,j}(\; \aTuple{s_1, \dots , s_n} \;) & = & \aTuple{s_j, s_j, s_j}
\end{eqnarray}

\noindent
The weight of the $n$-tuple $s\tapnum{n}$ is not modified by the projection
(if we consider $s\tapnum{n}$ not as a member of a relation).

The projection of an $n$-tape relation is the projection of all its string tuples:
\begin{eqnarray}
  \Proj{j,k,\dots}(\relat\tapnum{n})	\DefAs
	\aSet{ v\tapnum{m} \;\;|\;\;
		\exists s\tapnum{n}\!\in\!\relat\tapnum{n}
		\logAnd \Proj{j,k,\dots}(s\tapnum{n})\!=\!v\tapnum{m} }
\end{eqnarray}

\noindent
The weight of each
$v\tapnum{m}\!\in\!\Proj{j,k,\dots}(\relat\tapnum{n})$
is the collection (semiring sum) of the weights of each
$s\tapnum{n}\!\in\!\relat\tapnum{n}$
leading, when projected, to $v\tapnum{m}$:
\begin{equation}
  w( v\tapnum{m} )	\DefAs	\spc{-2ex}
	\srBigPlus_{ s\tapnum{n} \;|\; \Proj{j,k,\dots}(s\tapnum{n}) = v\tapnum{m} }
  \spc{-2ex}	w( s\tapnum{n} )
\end{equation}

The {\it complementary projection}\/, $\CProj{j,k,\dots}(s\tapnum{n})$,
of a string $n$-tuple $s\tapnum{n}$ removes all those strings (i.e., tapes) of the tuple
that are specified by the indices $j,k,\dots \in\rangeL 1, n \rangeR$,
and preserves all other strings in their original order.\footnote{
Contrary to other authors,
we do not call $\CProj{}(\;)$ an {\it inverse projection}\/
because it is not the inverse of a projection in the sense:
$\alpha=\Proj{}(\beta)$ and $\beta=\Proj{}^{-1}(\alpha)$.
}
It is defined as
\begin{equation}
  \CProj{j,k,\dots}(\; \aTuple{s_1, \dots , s_n} \;)	\DefAs
	\spc{5mm} \aTuple{\dots , s_{j-1}, s_{j+1}, \dots , s_{k-1}, s_{k+1}, \dots}
\end{equation}

\noindent
Complementary projection indices can occur in any order, but only once.

The complementary projection of an $n$-tape relation
equals the complementary projection of all its string tuples:
\begin{eqnarray}
  \CProj{j,k,\dots}(\relat\tapnum{n})	\DefAs
	\aSet{ v\tapnum{m} \;\;|\;\;
		\exists s\tapnum{n}\!\in\!\relat\tapnum{n}
		\logAnd \CProj{j,k,\dots}(s\tapnum{n})\!=\!v\tapnum{m} }
\end{eqnarray}

\noindent
The weight of each $v\tapnum{m}\!\in\!\CProj{j,k,\dots}(\relat\tapnum{n})$
is the collection of the weights of each $s\tapnum{n}\!\in\!\relat\tapnum{n}$
leading, when complementary projected, to $v\tapnum{m}$~:
\begin{equation}
  w( v\tapnum{m} )	\DefAs	\spc{-2ex}
	\srBigPlus_{ s\tapnum{n} \;|\; \CProj{j,k,\dots}(s\tapnum{n}) = v\tapnum{m} }
  \spc{-2ex}	w( s\tapnum{n} )
	\label{eq:Op:CProjWeight}
\end{equation}


\RSubSec{Auto-Intersection
		\label{sec:Op:AutoInt}}

We define the {\it auto-intersection}\/ of a relation,
$\AInt{j,k}(\relat\tapnum{n})$, on the tapes $j$ and $k$
as the subset of $\relat\tapnum{n}$
that contains all $s\tapnum{n}$ with equal $s_j$ and $s_k$:
\begin{equation}
  \AInt{j,k}(\; \relat\tapnum{n} \;)	\DefAs
	\aSet{\; s\tapnum{n}\!\in\!\relat\tapnum{n} \;\;|\;\; s_j=s_k \;}
    \label{eq:Op:AutoInt:DefOnRel}
\end{equation}

\noindent
The weight of any $s\tapnum{n}\!\in\!\AInt{j,k}(\relat\tapnum{n})$ is not modified.

\begin{figure}[ht]  
\begin{center}
\begin{minipage}{45mm}
\includegraphics[scale=0.50,angle=0]{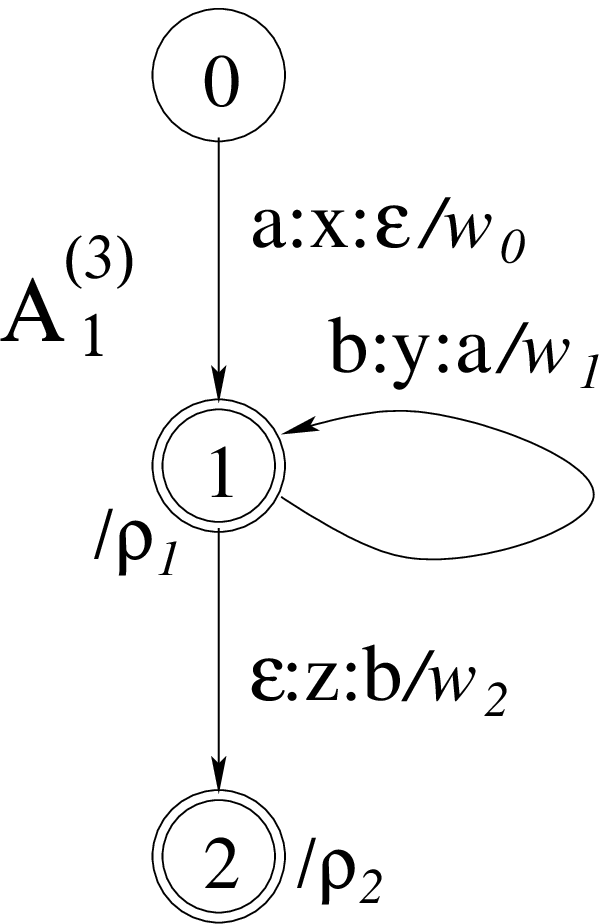}
\end{minipage}
\begin{minipage}{80mm}
\includegraphics[scale=0.50,angle=0]{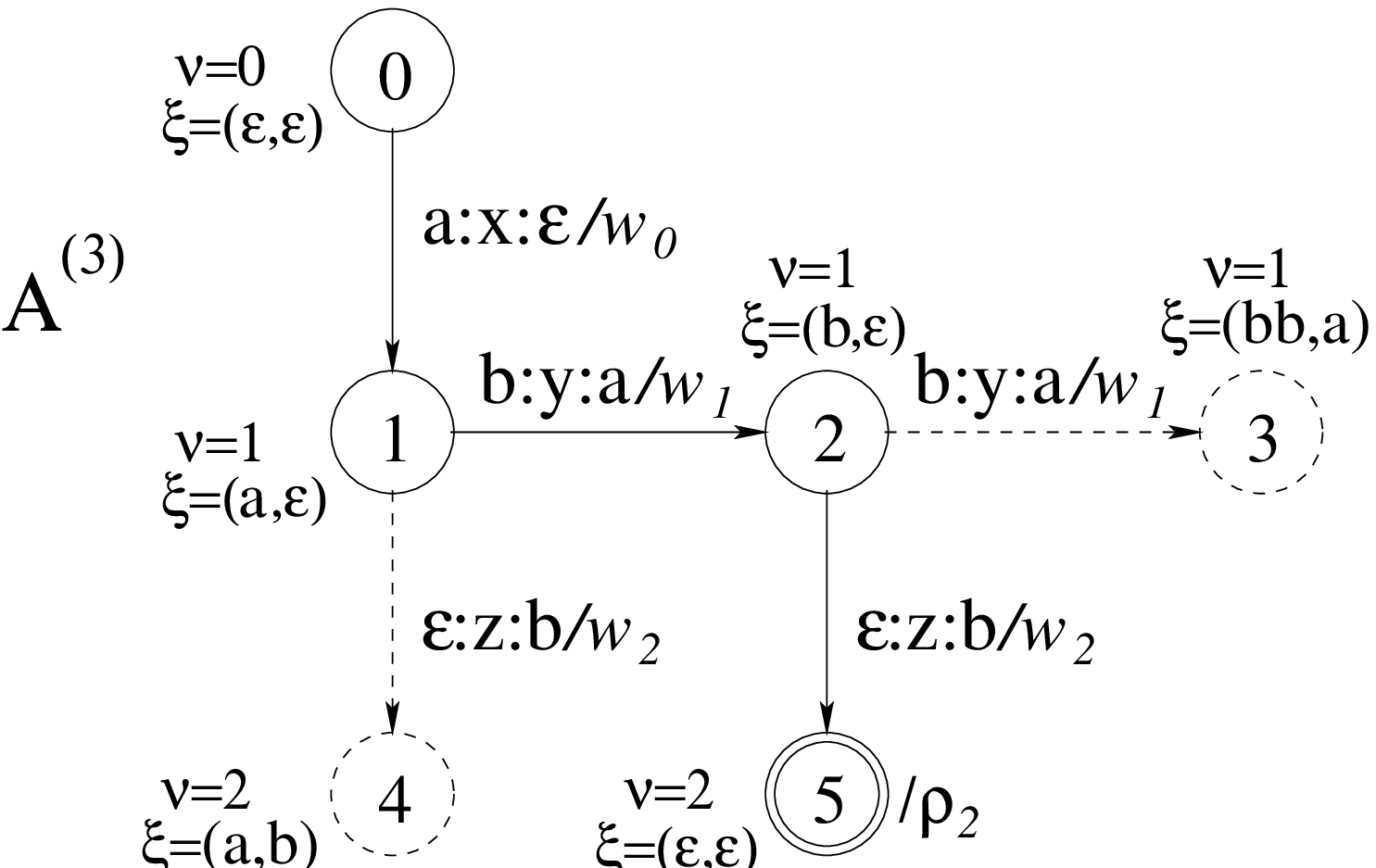}
\end{minipage}

\begin{minipage}{150mm}
\caption{A WMTA $A_1\tapnum{3}$ and its successfully constructed auto-intersection
	$A\tapnum{3}=\AInt{1,3}(A_1\tapnum{3})$.
	(Dashed parts are not constructed.)
	\label{fig:AInt:A11}}
\end{minipage}
\end{center}
\end{figure}

For example (Figure~\ref{fig:AInt:A11})
\begin{eqnarray}
  \relat_1\tapnum{3} & = &
	\aTuple{a,x,\eps} \;\aTuple{b,y,a}^* \;\aTuple{\eps,z,b} \;\;=\;\;
	\aSet{\; \aTuple{a b^k , x y^k z , a^k b} \;\;|\;\; k\!\in\!\srSetN \;}		\\
  \AInt{1,3}( \relat_1\tapnum{3} ) & = &
	\aSet{\; \aTuple{a b^1 , x y^1 z , a^1 b} \;}
\end{eqnarray}

Auto-intersection of regular $n$-tape relations is not necessarily regular.
For example (Figure~\ref{fig:AInt:A13})
\begin{eqnarray}
  \relat_2\tapnum{3} & = &
	\aTuple{a,\eps,x}^* \;\aTuple{a,a,y} \;\aTuple{\eps,a,z}^* \;\;=\;\;
	\aSet{\; \aTuple{a^k a , a a^h , x^k y z^h} \;\;|\;\; k,h\!\in\!\srSetN \;}
		\label{eq:Op:AutoInt:Exm2}		\\
  \AInt{1,2}( \relat_2\tapnum{3} ) & = &
	\aSet{\; \aTuple{a^k a , a a^k , x^k y z^k} \;\;|\;\; k\!\in\!\srSetN \;}
\end{eqnarray}

\noindent
The result is not regular because $x^k y z^k$ is not regular.


\RSubSec{Multi-Tape and Single-Tape Intersection
		\label{sec:Op:MlSgInt}}

The multi-tape intersection of two multi-tape relations, $\relat_1\tapnum{n}$ and $\relat_2\tapnum{m}$, 
uses $r$ tapes in each relation, and intersects them pair-wise. 
The operation pairs each string tuple $s\tapnum{n}\!\in\!\relat_1\tapnum{n}$ 
with each string tuple $v\tapnum{m}\!\in\!\relat_2\tapnum{m}$ 
iff $s_{j_i}\!=\!v_{k_i}$ with $j_i\!\in\! \rangeL 1,n \rangeR , k_i\!\in\! \rangeL 1,m \rangeR$
for all $i\!\in\! \rangeL 1,r \rangeR$.
Multi-tape intersection is defined as:
\begin{eqnarray}
  \relat_1\tapnum{n} \TTTInt{j_1,k_1}{\dots}{j_r,k_r} \relat_2\tapnum{m}
	\;\;=\;\;  \relat\tapnum{n+m-r}		\spc{50ex} \\
	\DefAs
	\{ u\tapnum{n+m-r} \;\;|\;\;
		\exists s\tapnum{n}\!\in\!\relat_1\tapnum{n} ,
		\exists v\tapnum{m}\!\in\!\relat_2\tapnum{m} ,\;
		s_{j_i}\!=\!v_{k_i} ,\;
		j_i\!\in\!\rangeL 1,n \rangeR , k_i\!\in\!\rangeL 1,m \rangeR ,
		\forall i\!\in\! \rangeL 1,r \rangeR		\nonumber \\
		u\tapnum{n+m-r}=\CProj{n+k_1,\dots,n+k_r}(s\tapnum{n}{\T}v\tapnum{m})\}	\nonumber
	\spc{25ex}
\end{eqnarray}

\noindent
All tapes $k_i$ of $\relat_2\tapnum{m}$ that have directly participated in the intersection 
are afterwards equal to the tapes $j_i$ of $\relat_1\tapnum{n}$, and are removed.
All tapes $j_i$ are kept for possible reuse by subsequent operations.
All other tapes of both relations are preserved without modification.

The weight of each $u\tapnum{n+m-r}\in\relat\tapnum{n+m-r}$ is
\begin{equation}
  w(\; u\tapnum{n+m-r} \;)	\;\;=\;\;
	w(s\tapnum{n}) \;\srTimes\; w(v\tapnum{m})
\end{equation}

\noindent
This weight follows only from pairing (Eq.~\ref{eq:Op:PairingWeight}).
It is not influenced by complementary projection (Eq.~\ref{eq:Op:CProjWeight})
because any two $u\tapnum{n+m}\!=\!s\tapnum{n}{\T}v\tapnum{m}$
that differ in $v_{k_i}$ also differ in $s_{j_i}$,
and hence cannot become equal when the $v_{k_i}$ are removed.

The multi-tape intersection of two relations,
$\relat_1\tapnum{n}$ and $\relat_2\tapnum{m}$, can be compiled by
\begin{equation}
  \relat_1\tapnum{n} \TTTInt{j_1,k_1}{\dots}{j_r,k_r} \relat_2\tapnum{m}  \;\; = \;\;
	\CProj{n+k_1,\dots,n+k_r}\left(\;
		\AInt{j_r,n+k_r}(\;
		\cdots	\;
		    \AInt{j_1,n+k_1}(\;
			\relat_1\tapnum{n} \!\CrPr\! \relat_2\tapnum{m}
		    \;)\;
		\cdots
	    \;)\;
	\right)
  \label{eq:Op:MltInt:proc}
\end{equation}

\noindent
as can been seen from
\begin{eqnarray}
  \relat_1\tapnum{n} \CrPr \relat_2\tapnum{m}	& = &
	\aSet{\; s\tapnum{n}{\T}v\tapnum{m} \;\;|\;\;
		s\tapnum{n}\in\relat_1\tapnum{n}, v\tapnum{m}\in\relat_2\tapnum{m} \;}	\\
  \AInt{j_1,n+k_1}( \relat_1\tapnum{n} \CrPr \relat_2\tapnum{m} )	& = &
	\aSet{\; s\tapnum{n}{\T}v\tapnum{m} \;\;|\;\;
		\exists s\tapnum{n}\in\relat_1\tapnum{n}, \exists v\tapnum{m}\in\relat_2\tapnum{m},
		s_{j_1}=v_{k_1} \;}		\\
	& \spc{-2ex}{\it etc.}\spc{-2ex} &		\nonumber
\end{eqnarray}

Multi-tape intersection is a generalization of classical intersection of transducers
which is known to be not necessarily regular
\cite{rabin+scott:1959}~:
\begin{equation}
  A_1\tapnum{2} \fsIntersect A_2\tapnum{2} 	\;\; = \;\;
  A_1\tapnum{2} \TTInt{1,1}{2,2} A_2\tapnum{2} 	\;\; = \;\;
  \CProj{3,4}\left(\; \AInt{2,4}(\; \AInt{1,3}(\; A_1\tapnum{2} \CrPr A_2\tapnum{2} \;) \;) \;\right)
\end{equation}

\noindent
Consequently, multi-tape intersection has the same property.
In our approach this results from the potential non-regularity of auto-intersection
(Eq.~\ref{eq:Op:MltInt:proc}).		\\


We speak about {\it single-tape intersection}\/ if only one tape is used in each relation ($r\!=\!1$).
A well-known special case is the intersection of two acceptors (1-tape automata)
leading to an acceptor
\begin{equation}
  A_1\tapnum{1} \fsIntersect A_2\tapnum{1} \;\; = \;\;
  A_1\tapnum{1} \TInt{1,1} A_2\tapnum{1} \;\; = \;\;
  \CProj{2}\left(\; \AInt{1,2}(\; A_1\tapnum{1} \CrPr A_2\tapnum{1} \;) \;\right)
\end{equation}

\noindent
and yielding the relation
\begin{eqnarray}
  \relat\left(\; A_1\tapnum{1} \fsIntersect A_2\tapnum{1} \;\right)  & = &
	\aSet{\; s \;\;|\;\; s\in\relat(A_1)\,,\; s\in\relat(A_2) \;}  \\
  w(s)  & = &  w_{A_1}(s) \srTimes w_{A_2}(s)
\end{eqnarray}

Another well-known special case is the composition
of two transducers (2-tape automata) leading to a transducer.
Here, we need, however, an additional complementary projection:\footnote{
Composition of transducers $T_i$ is expressed either by the $\diamond$
or the $\circ$ operator.
However, $T_1 \diamond T_2$ equals $T_2 \circ T_1$
which corresponds to ${\rm T}_2(\,{\rm T}_1(\;\;)\,)$ in functional notation
\cite{birkhoff+bartee:1970}.
}
\begin{equation}
  A_1\tapnum{2} \fsCompose A_2\tapnum{2} \;\; = \;\;
  \CProj{2}(\; A_1\tapnum{2} \TInt{2,1} A_2\tapnum{2} \;) \;\; = \;\;
  \CProj{2,3}\left(\; \AInt{2,3}(\; A_1\tapnum{2} \CrPr A_2\tapnum{2} \;) \;\right)
\end{equation}

\noindent
It yields the relation:
\begin{eqnarray}
  \relat\left( A_1\tapnum{2} \fsCompose A_2\tapnum{2} \;\right)  & = &
	\aSet{ u\tapnum{2} \;|\;
		\exists s\tapnum{2}\!\!\in\!\!\relat(A_1\tapnum{2}) ,
		\exists v\tapnum{2}\!\!\in\!\!\relat(A_2\tapnum{2}) ,
		s_2\!=\!v_1 ,
		u\tapnum{2}\!=\!\CProj{2,3}(s\tapnum{2}{\T}v\tapnum{2}) }	\spc{4ex} \\
  w(u\tapnum{2})  & = & \spc{-2ex}
	\srBigPlus_{ s\tapnum{2}, v\tapnum{2} \;|\; u_1=s_1, s_2=v_1 , v_2=u_2 }
  \spc{-2ex}	w_{A_1}(s\tapnum{2}) \;\srTimes\; w_{A_2}(v\tapnum{2})
\end{eqnarray}

Multi-tape and single-tape intersection are neither associative nor commutative,
except for special cases with $r=n=m$,
such as the above intersection of acceptors and transducers.


\RSubSec{Transduction
		\label{sec:Op:Transduct}}

A WMTA, $A\tapnum{n}$,
can be used as a transducer having $r$ input tapes, $j_1$ to $j_r$,
and $x$ output tapes, $k_1$ to $k_x$,
which do not have to be consecutive or disjoint.

To apply $A\tapnum{n}$ to a weighted $r$-tuple of input strings,
the tuple $s\tapnum{r}$ is converted into an input WMTA, $I\tapnum{r}$,
having one single path labeled with $s\tapnum{r}$ and weighted with $w(s\tapnum{r})$.
An output WMTA, $O\tapnum{x}$,
whose relation contains all weighted $x$-tuples of output strings, $v\tapnum{x}$,
is then obtained through multitape-intersection and projection:
\begin{equation}
  O\tapnum{x}  \; = \;
	\Proj{k_1,\dots,k_x}(\;
		A\tapnum{n}  \TTTInt{j_1,1}{\cdots}{j_r,r}  I\tapnum{r}
	  \;)
\end{equation}


\renewcommand{\RSec}[1]{\section{#1}}
\renewcommand{\RSubSec}[1]{\subsection{#1}}
\renewcommand{\RSubSubSec}[1]{\subsubsection{#1}}

\RSec{Example of Classical Transducer Intersection
			\label{sec:Exm:2tapeIntersect}}

The following example of classical transducer intersection
of $A_1\tapnum{2}$ and $A_2\tapnum{2}$ is regular:\footnote{
For sake of space and clarity we represent all regular expressions
in this section in a special form
where each tape appears on a different row
and symbols of the same transition are vertically aligned.
Note that it is not a matrix representation.
More conventionally $A_1\tapnum{2}$ could be written as~
$
\aTuple{a,\eps} \aTuple{b,A} \;
\left(\; \aTuple{c,B} \aTuple{a,\eps} \aTuple{b,C} \;\right)^* \;
\aTuple{\eps,A} \aTuple{\eps,B} \aTuple{\eps,C} \aTuple{c,\eps} \aTuple{\eps,A}
$.
}
\vspace{1ex}

\STTSep{1ex}
$
\begin{StringTupleTable}{2}
  a     & b     \\
  \epN  & A     \\
\end{StringTupleTable}
\;\left(
\begin{StringTupleTable}{3}
  c     & a     & b     \\
  B     & \epN  & C     \\
\end{StringTupleTable}
\right)\bigExp{*}
\begin{StringTupleTable}{5}
  \epN  & \epN  & \epN  & c     & \epN  \\
  A     & B     & C     & \epN  & A     \\
\end{StringTupleTable}
\spc{4ex} \TTInt{1,1}{2,2} \spc{4ex}
\begin{StringTupleTable}{1}
  \epN  \\
  A     \\
\end{StringTupleTable}
\;\left(
\begin{StringTupleTable}{4}
  a     & b     & \epN  & c     \\
  B     & \epN  & C     & A     \\
\end{StringTupleTable}
\right)\bigExp{*}
$

\noindent
It has one theoretical solution which is
\vspace{1ex}

\STTSep{0.7ex}
$
\begin{StringTupleTable}{2}
  a     & b     \\
  \epN  & A     \\
\end{StringTupleTable}
\;\left(
\begin{StringTupleTable}{3}
  c     & a     & b     \\
  B     & \epN  & C     \\
\end{StringTupleTable}
\right)\bigExp{1}
\begin{StringTupleTable}{5}
  \epN  & \epN  & \epN  & c     & \epN  \\
  A     & B     & C     & \epN  & A     \\
\end{StringTupleTable}
\spc{1ex} = \spc{1ex}
\begin{StringTupleTable}{7}
  a     & b     & c     & a     & b     & c     & \epN \\
  A     & B     & C     & A     & B     & C     & A    \\
\end{StringTupleTable}
\spc{1ex} = \spc{1ex}
\begin{StringTupleTable}{1}
  \epN  \\
  A     \\
\end{StringTupleTable}
\;\left(
\begin{StringTupleTable}{4}
  a     & b     & \epN  & c     \\
  B     & \epN  & C     & A     \\
\end{StringTupleTable}
\right)\bigExp{2}
$	\\

This solution cannot be compiled with any of the above mentioned previous approaches
(Section~\ref{sec:Previous}).
It cannot be enabled by any pre-transformation of the WMTAs
that does not change their relations,
$\relat(A_1\tapnum{2})$ and $\relat(A_2\tapnum{2})$.
All above mentioned approaches do not exceed the following alternatives.


\RSubSec{First Failing Alternative}

One can start by typing all symbols (and $\eps$) with respect to the tapes,
to make the alphabets of different tapes disjoint
(which can be omitted for symbols occurring on one tape only)~:
\vspace{1ex}

\STTSep{1ex}
$
\begin{StringTupleTable}{2}
  a     & b     \\
  $\eps_2$  & A     \\
\end{StringTupleTable}
\;\left(
\begin{StringTupleTable}{3}
  c     & a     & b     \\
  B     & $\eps_2$  & C     \\
\end{StringTupleTable}
\right)\bigExp{*}
\begin{StringTupleTable}{5}
  $\eps_1$  & $\eps_1$  & $\eps_1$  & c     & $\eps_1$  \\
  A     & B     & C     & $\eps_2$  & A     \\
\end{StringTupleTable}
\spc{4ex} \TTInt{1,1}{2,2} \spc{4ex}
\begin{StringTupleTable}{1}
  $\eps_1$  \\
  A     \\
\end{StringTupleTable}
\;\left(
\begin{StringTupleTable}{4}
  a     & b     & $\eps_1$  & c     \\
  B     & $\eps_2$  & C     & A     \\
\end{StringTupleTable}
\right)\bigExp{*}
$	\\

\noindent
Then, one converts $n$ tapes into $1$ tape,
such that each transition, labeled with $n$ symbols,
is transformed into a sequence of $n$ transitions, labeled with $1$ symbol each,
which is equivalent to Ganchev's approach \cite{ganchev+al:2003}~:
\vspace{1ex}

\STTSep{0.5ex}
$
\begin{StringTupleTable}{4}
  a     & $\eps_2$  & b     & A     \\
\end{StringTupleTable}
\;\left(
\begin{StringTupleTable}{6}
  c     & B     & a     & $\eps_2$  & b     & C     \\
\end{StringTupleTable}
\right)\bigExp{*}
\begin{StringTupleTable}{10}
  $\eps_1$  & A     & $\eps_1$  & B     & $\eps_1$  & C     & c     & $\eps_2$  & $\eps_1$  &  A     \\
\end{StringTupleTable}
\spc{2ex} \cap \spc{2ex}
\begin{StringTupleTable}{2}
  $\eps_1$  & A     \\
\end{StringTupleTable}
\;\left(
\begin{StringTupleTable}{8}
  a     & B     & b     & $\eps_2$  & $\eps_1$  & C     & c     & A     \\
\end{StringTupleTable}
\right)\bigExp{*}
$	\\

\noindent
After these transformations,
it is not possible to obtain the above theoretical solution
by means of classical intersection of 1-tape automata,
even not after $\eps$-removal:
\vspace{1ex}

\STTSep{0.8ex}
$
\begin{StringTupleTable}{3}
  a     & b     & A     \\
\end{StringTupleTable}
\;\left(
\begin{StringTupleTable}{5}
  c     & B     & a     & b     & C     \\
\end{StringTupleTable}
\right)\bigExp{*}
\begin{StringTupleTable}{5}
  A     & B     & C     & c     &  A     \\
\end{StringTupleTable}
\spc{4ex} \cap \spc{4ex}
\begin{StringTupleTable}{1}
  A     \\
\end{StringTupleTable}
\;\left(
\begin{StringTupleTable}{6}
  a     & B     & b     & C     & c     & A     \\
\end{StringTupleTable}
\right)\bigExp{*}
$


\RSubSec{Second Failing Alternative}

Alternatively,
one could start with synchronizing the WMTAs.
This is not possible across a whole WMTA, but only within ``limited sections'':
in our example this means before, inside, and after the cycles:
\vspace{1ex}

\STTSep{1ex}
$
\begin{StringTupleTable}{2}
  a     & b     \\
  A     & \epN  \\
\end{StringTupleTable}
\;\left(
\begin{StringTupleTable}{3}
  c     & a     & b     \\
  B     & C     & \epN  \\
\end{StringTupleTable}
\right)\bigExp{*}
\begin{StringTupleTable}{4}
  c     & \epN  & \epN  & \epN  \\
  A     & B     & C     & A     \\
\end{StringTupleTable}
\spc{4ex} \TTInt{1,1}{2,2} \spc{4ex}
\begin{StringTupleTable}{1}
  \epN  \\
  A     \\
\end{StringTupleTable}
\;\left(
\begin{StringTupleTable}{3}
  a     & b     & c     \\
  B     & C     & A     \\
\end{StringTupleTable}
\right)\bigExp{*}
$	\\

\noindent
Then, one can proceed as before by first typing the symbols with respect to the tapes
\vspace{1ex}

$
\begin{StringTupleTable}{2}
  a     & b     \\
  A     & $\eps_2$  \\
\end{StringTupleTable}
\;\left(
\begin{StringTupleTable}{3}
  c     & a     & b     \\
  B     & C     & $\eps_2$  \\
\end{StringTupleTable}
\right)\bigExp{*}
\begin{StringTupleTable}{4}
  c     & $\eps_1$  & $\eps_1$  & $\eps_1$  \\
  A     & B     & C     & A     \\
\end{StringTupleTable}
\spc{4ex} \TTInt{1,1}{2,2} \spc{4ex}
\begin{StringTupleTable}{1}
  $\eps_1$  \\
  A     \\
\end{StringTupleTable}
\;\left(
\begin{StringTupleTable}{3}
  a     & b     & c     \\
  B     & C     & A     \\
\end{StringTupleTable}
\right)\bigExp{*}
$	\\

\noindent
and then transforming $n$ tapes into $1$ tape
\vspace{1ex}

\STTSep{0.5ex}
$
\begin{StringTupleTable}{4}
  a     & A     & b     & $\eps_2$  \\
\end{StringTupleTable}
\;\left(
\begin{StringTupleTable}{6}
  c     & B     & a     & C     & b     & $\eps_2$  \\
\end{StringTupleTable}
\right)\bigExp{*}
\begin{StringTupleTable}{8}
  c     & A     & $\eps_1$  & B     & $\eps_1$  & C     & $\eps_1$  & A     \\
\end{StringTupleTable}
\spc{4ex} \cap \spc{4ex}
\begin{StringTupleTable}{2}
  $\eps_1$  & A     \\
\end{StringTupleTable}
\;\left(
\begin{StringTupleTable}{6}
  a     & B     & b     & C     & c     & A     \\
\end{StringTupleTable}
\right)\bigExp{*}
$	\\

\noindent
The solution cannot be compiled with this alternative either,
even not after $\eps$-removal:
\vspace{1ex}

\STTSep{0.8ex}
$
\begin{StringTupleTable}{3}
  a     & A     & b     \\
\end{StringTupleTable}
\;\left(
\begin{StringTupleTable}{5}
  c     & B     & a     & C     & b     \\
\end{StringTupleTable}
\right)\bigExp{*}
\begin{StringTupleTable}{5}
  c     & A     & B     & C     & A     \\
\end{StringTupleTable}
\spc{4ex} \cap \spc{4ex}
\begin{StringTupleTable}{1}
  A     \\
\end{StringTupleTable}
\;\left(
\begin{StringTupleTable}{6}
  a     & B     & b     & C     & c     & A     \\
\end{StringTupleTable}
\right)\bigExp{*}
$


\RSubSec{Solution with Our Approach}

To compile multi-tape intersection according to the above procedure
(Eq.~\ref{eq:Op:MltInt:proc})
\begin{equation}
  A\tapnum{2}
	\;=\;  A_1\tapnum{2}  \TTInt{1,1}{2,2}  A_2\tapnum{2}
	\;=\;  \CProj{3,4}(\;\AInt{2,4}(\;\AInt{1,3}(\; A_1\tapnum{2} \CrPr A_2\tapnum{2} \;)\;)\;)  \\
    \nonumber
\end{equation}

\noindent
we proceed in 3 steps.
First, we compile
$B_1\tapnum{4} = \AInt{1,3}( A_1\tapnum{2} \CrPr A_2\tapnum{2} )$
in one single step with an algorithm that follows the principle of transducer composition
and simulates the behaviour of Mohri's $\eps$-filter
(Section~\ref{sec:Alg:SgInt}).\footnote{
Composition with $\eps$-filter has been shown to work on arbitrary transducers
\cite{mohri+al:1998}.
}
For the above example, we obtain
\vspace{1ex}

\STTSep{1.5ex}
$
\begin{StringTupleTable}{3}
  \epN  & a     & b     \\
  \epN  & \epN  & A     \\
  \epN  & a     & b     \\
  A     & B     & \epN  \\
\end{StringTupleTable}
\;\left(
\begin{StringTupleTable}{4}
  \epN  & c     & a     & b     \\
  \epN  & B     & \epN  & C     \\
  \epN  & c     & a     & b     \\
  C     & A     & B     & \epN  \\
\end{StringTupleTable}
\right)\bigExp{*}
\begin{StringTupleTable}{5}
  \epN  & \epN  & \epN  & c     & \epN  \\
  A     & B     & C     & \epN  & A     \\
  \epN  & \epN  & \epN  & c     & \epN  \\
  C     & \epN  & \epN  & A     & \epN  \\
\end{StringTupleTable}
$	\\
\vspace{1ex}

\noindent
Next, we compile
$B_2\tapnum{4} = \AInt{2,4}( B_1\tapnum{4} )$
using our auto-intersection algorithm
(Section~\ref{sec:Alg:AutoInt})
\vspace{1ex}

$
\begin{StringTupleTable}{3}
  \epN  & a     & b     \\
  \epN  & \epN  & A     \\
  \epN  & a     & b     \\
  A     & B     & \epN  \\
\end{StringTupleTable}
\;\left(
\begin{StringTupleTable}{4}
  \epN  & c     & a     & b     \\
  \epN  & B     & \epN  & C     \\
  \epN  & c     & a     & b     \\
  C     & A     & B     & \epN  \\
\end{StringTupleTable}
\right)\bigExp{1}
\begin{StringTupleTable}{5}
  \epN  & \epN  & \epN  & c     & \epN  \\
  A     & B     & C     & \epN  & A     \\
  \epN  & \epN  & \epN  & c     & \epN  \\
  C     & \epN  & \epN  & A     & \epN  \\
\end{StringTupleTable}
$	\\
\vspace{1ex}

\pagebreak  

\noindent
and finally,
$A\tapnum{2} = \CProj{3,4}(\; B_2\tapnum{4} \;)$
with a simple algorithm for complementary projection:
\vspace{1ex}

$
\begin{StringTupleTable}{3}
  \epN  & a     & b     \\
  \epN  & \epN  & A     \\
\end{StringTupleTable}
\;\left(
\begin{StringTupleTable}{4}
  \epN  & c     & a     & b     \\
  \epN  & B     & \epN  & C     \\
\end{StringTupleTable}
\right)\bigExp{1}
\begin{StringTupleTable}{5}
  \epN  & \epN  & \epN  & c     & \epN  \\
  A     & B     & C     & \epN  & A     \\
\end{StringTupleTable}
$	\\

\noindent
This final result equals the theoretical solution.


\renewcommand{\RSec}[1]{\section{#1}}
\renewcommand{\RSubSec}[1]{\subsection{#1}}
\renewcommand{\RSubSubSec}[1]{\subsubsection{#1}}

\RSec{Algorithms}

In this section we propose and recall algorithms for the above defined operations on WMTAs:
cross-product, auto-intersection, single-tape and multi-tape intersection.
By convention, our WMTAs have only one initial state $i\!\in\! I$,
without loss of generality,
since for any WMTA with multiple initial states there exists a WMTA
with a single initial state accepting the same relation.

We will use the following variables and definitions.
The variables $\nu[q]$, $\mu[q]$, etc.
serve for assigning temporarily additional data to a state $q$. \\

\noindent
\begin{tabular}{p{12mm} p{34mm} p{102mm}}
$A_j$	& $=\;$\scalebox{0.8 1.0}{$\aTuple{\Sigma_j,Q_j,i_j,F_j,E_j,\srK_j}$}	&
	Original weighted automaton
	from which we will construct a new weighted automaton $A$	\\
$A$	& $=\aTuple{\Sigma,Q,i,F,E,\srK}$		&
	New weighted automaton resulting from a construction	\\
$\nu[q]$ & $=q_1$	&
	State $q_1$ of an original automaton $A_1$
	assigned to a state $q$ of a new automaton $A$	\\
$\mu[q]$ & $=(q_1, q_2)$	&
	pair of states $(q_1, q_2)$ of two original automata, $A_1$ and $A_2$,
	assigned to a state $q$ of a new automaton $A$	\\
$\vartheta[q]$ & $=(q_1, q_2, q_\eps)$	&
	triple of states belonging to the two original automata, $A_1$ and $A_2$,
	and to a simulated filter automaton, $A_\eps$, respectively;
	assigned to a state $q$ of a new automaton $A$	\\
$\xi[q]$ & $=(s, u)$	&
	Pair of ``leftover'' substrings $(s, u)$
	assigned to a state $q$ of a new automaton $A$	\\
$\delta(s, u)$	& $=|s|\!-\!|u|$	&
	Delay between two string (or leftover substrings) $s$ and $u$.
	For example: $\delta(\xi[q])$ also written as $\delta(q)$		\\
$\chi[q]$ & $=(\chi_1, \chi_2)$	&
	Pair of integers assigned to a state $q$,
	expressing the lengths of two strings $s$ and $u$
	on different tapes of the same path ending at $q$		\\
\FCT{lcp}{$s, s^\prime$}	&	&
	Longest common prefix of the strings $s$ and $s^\prime$  \\
$\lab_{j,k,\dots}(e)$	& $=\Proj{j,k,\dots}(\; \lab(e) \;)$	&
	Short-hand notation for the projection of the label of $e$  \\
\end{tabular}


\renewcommand{\RSec}[1]{\subsection{#1}}
\renewcommand{\RSubSec}[1]{\subsubsection{#1}}
\renewcommand{\RSubSubSec}[1]{\subsubsection{SUB -- #1}}


\RSec{Cross Product
	\label{sec:Alg:CrProd}}

We describe two alternative algorithms to compile the cross product of two WMTAs,
$A_1\tapnum{n}$ and $A_2\tapnum{m}$.
The second algorithm is almost identical to classical algorithms for crossproduct of automata.
Nevertheless, we recall it to make this report more complete and self-standing.

\RSubSec{Conditions}

Both algorithms require the semirings of the two original automata,
$A_1\tapnum{n}$ and $A_2\tapnum{m}$, to be equal ($\srK_1\!=\!\srK_2$).
The second algorithm requires the common semiring
$\srK\!=\!\srK_1\!=\!\srK_2$ to be commutative.

\RSubSec{Algorithms}

{\bf Cross product through path concatenation:}~
The first algorithm pairs
the label of each transition $e_1\!\in\!E_1$ with $\eps\tapnum{m}$
(producing $\lab(e_1)\!\T\!\eps\tapnum{m}$),
and the label of each transition $e_2\!\in\!E_2$ with $\eps\tapnum{n}$
(producing $\eps\tapnum{n}\!\T\!\lab(e_2)$),
and then concatenates $A_1\tapnum{n+m}$ with $A_2\tapnum{n+m}$.
We will refer to it as \FUNCT{CrossPC}{$A_1, A_2$}
where the suffix {\it PC} stands for {\it path concatenation}\/.

\input{CrossPC.pc}

We start with a WMTA $A$ that is equipped with the union of the alphabets,
the union of the state sets, and the union of the transition sets of $A_1$ and $A_2$.
The initial state of $A$ equals that $A_1$, its set of final states equals that of $A_2$,
and its semiring equals those of $A_1$ and $A_2$
(Line~\ref{pc:CrossPC:L1}).
First, we (post-) pair the labels of all transitions originally coming from $A_1$
with $\eps\tapnum{m}$,
and (pre-) pair the labels of all transition from $A_2$ with $\eps\tapnum{n}$.
Then, we connect all final states of $A_1$ with the initial state of $A_2$
through $\eps\tapnum{n+m}$-transitions,
as is usually done in the concatenation of automata.

The disadvantages of this algorithm are that the paths of $A$ become longer
than in the second algorithm below
and that each transition of $A$ is partially labeled with $\eps$,
which may increase the running time of subsequently applied operations.

To adapt this algorithm to non-weighted MTAs,
one has to remove the weight from Line~\ref{pc:CrossPC:L7}
and replace Line~\ref{pc:CrossPC:L8} with:
$Final(q)$ \SETTO {\it false}.		\\

{\bf Cross product through path alignment:}~
The second algorithm pairs each string tuple of $A_1\tapnum{n}$
with each string tuple of $A_2\tapnum{m}$, following the definition
(Eq.~\ref{eq:Op:CrProd}).
The algorithm actually pairs each path $\path_1$ of $A_1\tapnum{n}$
with each path $\path_2$ of $A_2\tapnum{m}$ transition-wise,
and appends $\eps$-transitions to the shorter of two paired paths,
so that both have equal length.
We will refer to this algorithm as \FUNCT{CrossPA}{$A_1, A_2$}
where the suffix {\it PA} stands for {\it path alignment}\/.

We start with a WMTA $A$ whose alphabet is the union of the alphabets of $A_1$ and $A_2$,
whose semiring equals those of $A_1$ and $A_2$,
and that is otherwise empty
(Line~\ref{pc:CrossPA:L1}).
First, we create the initial state $i$ of $A$ from the initial states of $A_1$ and $A_2$,
and push $i$ onto the stack
(Lines~\ref{pc:CrossPA:L3}, \ref{pc:CrossPA:L27}--\ref{pc:CrossPA:L33}).
While the stack is not empty,
we take states $q$ from it and access the states $q_1$ and $q_2$
that are assigned to $q$ through $\mu[q]$
(Lines~\ref{pc:CrossPA:L5},~\ref{pc:CrossPA:L6}).

\input{CrossPA.pc}

If both $q_1$ and $q_2$ are defined $(\not=\!\Null)$,
we pair each outgoing transition $e_1$ of $q_1$
with each outgoing transition $e_2$ of $q_2$
(Lines~\ref{pc:CrossPA:L7}--\ref{pc:CrossPA:L9}),
and create a transition in $A$
(Line~\ref{pc:CrossPA:L13})
whose label is the pair $\lab(e_1) \paired \lab(e_2)$ and
whose target $q^\prime$ corresponds to the tuple of targets $(n(e_1),n(e_2))$
(Line~\ref{pc:CrossPA:L10}).
If $q^\prime$ does not exist yet,
it is created and pushed onto the stack
(Lines~\ref{pc:CrossPA:L27}--\ref{pc:CrossPA:L33}).

If we encounter a final state $q_1$ (with $\wgtFin(q_1)\!\not=\!\srZero$) in $A_1$,
we follow the path beyond $q_1$ on an $\eps$-transition
that exists only ``virtually'' but not ``physically'' in $A_1$
(Lines~\ref{pc:CrossPA:L14},~\ref{pc:CrossPA:L15}).
The target of the resulting transition in $A$ corresponds to the tuple
of targets $(n(e_1), n(e_2))$ with $n(e_1)$ being undefined $(=\!\Null)$
because $e_1$ does not exist physically 
(Line~\ref{pc:CrossPA:L16}).
If we encounter a final state $q_2$ (with $\wgtFin(q_2)\!\not=\!\srZero$) in $A_2$,
we proceed similarly
(Lines~\ref{pc:CrossPA:L20}--\ref{pc:CrossPA:L25}).

The final weight of an undefined state $q=\Null$ is assumed to be $\srOne$~:
$\wgtFin(\Null)=\srOne$~.

To adapt this algorithm to non-weighted MTAs,
one has to remove the weights from the
Lines~\ref{pc:CrossPA:L13},~\ref{pc:CrossPA:L19}, and~\ref{pc:CrossPA:L25},
and replace Line~\ref{pc:CrossPA:L30} with:
$Final(q)$ \SETTO $Final(q_1) \logAnd Final(q_2)$.


\RSec{Auto-Intersection
	\label{sec:Alg:AutoInt}}

We propose an algorithm that attempts to constructs
the auto-intersection $A\tapnum{n}$ of a WMTA $A_1\tapnum{n}$.
Our approach has some minor similarity with synchronization algorithms for transducers
\cite{frougny+sakarovitch:1993,mohri:2003}~:
it uses the concept of delay between two tapes and assigns leftover-strings to states
(see above).

In the context of our approach, we understand by {\it construction}\/
the compilation of reachable states $q$ and transitions $e\tapnum{n}$ of $A\tapnum{n}$,
such that the absolute value of the delay $\delta(q)$, regarding tape $j$ and $k$,
does not exceed a limit $\delta_{\rm max2}$ at any state $q$, i.e.:
$ \forall q :  |\delta(q)| \leq \delta_{\rm max2}  \logAnd  q {\sf ~reachable} $.
The limit $\delta_{\rm max2}$ is imposed, i.e.,
any state whose delay would exceed it is not constructed.

We distinguish two cases.
In the first case, the delay of none of the reachable and coreachable states
exceeds a limit $\delta_{\rm max}$ ~(with $\delta_{\rm max}\leq\delta_{\rm max2}$), i.e.:
$ \not\!\!\exists q : \delta_{\rm max} < |\delta(q)| \leq \delta_{\rm max2}
\logAnd  q {\sf ~reachable} \logAnd  q {\sf ~coreachable} $.
We call it a construction with {\it bounded delay}\/
or a {\it successful}\/ construction
because it is guarantied to generate the attempted result $A\tapnum{n}=\AInt{j,k}(A_1\tapnum{n})$.
In this case the relation $\AInt{j,k}(A_1\tapnum{n})$ has bounded delay, too,
and is rational.\footnote{
A {\it rational}\/ relation is a weighted regular relation.
}
The limit $\delta_{\rm max}$ is not imposed, i.e.,
any state $q$ whose delay exceeds it would still be constructed
(which places the construction into the second case if $q$ becomes coreachable).

In the second case, the delay of reachable and coreachable states is potentially unbounded.
It exceeds $\delta_{\rm max}$, and would actually exceed any limit
if it was not (brute-force) delimited by $\delta_{\rm max2}$, i.e.:
$ \exists q : \delta_{\rm max} < |\delta(q)| \leq \delta_{\rm max2}
\logAnd  q {\sf ~reachable} \logAnd  q {\sf ~coreachable} $.
We call this a construction with {\it potentially unbounded delay}\/.
It is not successful,
and we cannot conclude on the correctness of the result $A\tapnum{n}$
and on the boundedness and rationality of the relation $\AInt{j,k}(A_1\tapnum{n})$.

We will first describe the algorithm
and then present some examples for further illustration.

\RSubSec{Algorithm}

Our algorithm starts with the compilation of the limits 
$\delta_{\rm max}$ and $\delta_{\rm max2}$, 
then proceeds with the construction of $A\tapnum{n}$,
and finally verifies the success of the construction,
according to the above conditions.	\\

{\bf Compilation of limits:}~
First, we traverse $A_1\tapnum{n}$ recursively,
without traversing any state more than once,
and record three values:
$\widehat\delta_{max}$, being the maximal delay at any state,
$\widehat\delta_{min}$, the minimal delay at any state, and
$\widehat\delta_{cyc}$, the maximal absolute value of the delay of any cycle
(Lines~\ref{pc:AutoInt:L403}, \ref{pc:AutoInt:L501}--\ref{pc:AutoInt:L510}).
To do so, we assign to each state $q_1$ of $A_1\tapnum{n}$ a variable
$\chi[q_1]\!=\!(\chi_1,\chi_2)$ with the above defined meaning.
The delay at a state $q_1$ is $\delta(q_1)\!=\!\chi_1\!-\!\chi_2$
(Lines~\ref{pc:AutoInt:L501},~\ref{pc:AutoInt:L502}).
The delay of a cycle on $q_1$ is the difference
between $\delta^\prime(q_1)$ at the end
and $\delta(q_1)$ at the beginning of the cycle
(Line~\ref{pc:AutoInt:L504}).

Then, we compile $\delta_{cyc}$,
the maximal absolute value of delay required to match any two cycles.
For example, let
$\relat(A_1\tapnum{2})=\left(\aSet{\aTuple{aa,\eps}}\cup\aSet{\aTuple{\eps,aaa}}\right)^*$,
encoded by two cycles.
To obtain a match between $\plab_1(\path)$ and $\plab_2(\path)$
of a path $\pi$ of $A\tapnum{2}\!\subseteq\!\AInt{1,2}(A_1\tapnum{2})$,
we have to traverse the first cycle 3 times and the second two times,
allowing for any permutation:
$
  A\tapnum{2} = (
	\aTuple{aa,\eps}^3 \aTuple{\eps,aaa}^2  \cup
	\aTuple{aa,\eps}^2 \aTuple{\eps,aaa}^2 \aTuple{aa,\eps}^1  \cup
	\ldots )^*
$.
This illustrates that in a match between any two cycles of $A_1\tapnum{n}$,
the absolute value of the delay does not exceed
$\delta_{cyc}=\widehat\delta_{\rm cyc}\!\cdot\!\FCT{max}{\,1 , \,\widehat\delta_{\rm cyc}\!-\!1\,}$
(Line~\ref{pc:AutoInt:L404}).

\input{AInt2.pc}

Next, we compile the first limit, $\delta_{\rm max}$,
that will not be exceeded by a construction with bounded delay.
In a match of two cycles this limit equals $\delta_{cyc}$,
and for any other match it is $\widehat\delta_{max}\!-\!\widehat\delta_{min}$.
In a construction with bounded delay,
the absolute value of the delay in $A\tapnum{n}$ does therefore not exceed
$\delta_{max}=\FCT{max}{\delta_{cyc}\;,\; \widehat\delta_{max}\!-\!\widehat\delta_{min}}$
(Line~\ref{pc:AutoInt:L405}).

Finally, we compile a second limit, $\delta_{\rm max2}$,
that allows us, in case of potentially unbounded delay,
to construct a larger $A\tapnum{n}$ than $\delta_{\rm max}$ does.
Unboundedness can only result from matching cycles in $A_1\tapnum{n}$.
To obtain a larger $A\tapnum{n}$,
with states whose delay exceeds $\delta_{\rm max}$,
we have to unroll the cycles of $A_1\tapnum{n}$ further
until we reach (at least) one more match between two cycles.
Therefore,
$\delta_{max2}=\delta_{max}\!+\!\delta_{cyc}$
(Line~\ref{pc:AutoInt:L405a}).		\\

{\bf Construction:}~
We start with a WMTA $A$ whose alphabet and semiring equal those of $A_1$
and that is otherwise empty
(Line~\ref{pc:AutoInt:L102}).
To each state $q$ that will be created in $A$,
we will assign two variables:
$\nu[q]\!=\!q_1$ indicating the corresponding state $q_1$ in $A_1$,
and $\xi[q]\!=\!(s, u)$ stating
the leftover string $s$ of tape $j$ (yet unmatched in tape $k$)
and the leftover string $u$ of tape $k$ (yet unmatched in tape $j$).

Then, we create an initial state $i$ in $A$ and push it onto the stack
(Lines~\ref{pc:AutoInt:L104}, \ref{pc:AutoInt:L301}--\ref{pc:AutoInt:L310}).
As long as the stack is not empty,
we take states $q$ from it and follow each of the outgoing transitions
$e_1\!\in\!E(q_1)$ of the corresponding state $q_1\!=\!\nu[q]$ in $A_1$
(Lines~\ref{pc:AutoInt:L105}--\ref{pc:AutoInt:L106}).
A transition $e_1$ in $A_1$ is represented as $e\!\in\!E(q)$ in $A$,
with the same label and weight.
To compile the leftover strings $\xi[q^\prime]\!=\!(s^\prime, u^\prime)$
of its target $q^\prime\!=\!n(e)$ in $A$,
we concatenate the leftover strings $\xi[q]\!=\!(s, u)$ of its source $q\!=\!p(e)$
with the $j$-th and $k$-th component of its label, $\lab_j(e_1)$ and $\lab_k(e_1)$,
and remove the longest common prefix of the resulting strings
$s\cdot\lab_j(e_1)$ and $u\cdot\lab_k(e_1)$
(Lines~\ref{pc:AutoInt:L107}, \ref{pc:AutoInt:L201}--\ref{pc:AutoInt:L204}).

\input{AInt1.pc}

If both leftover strings $s^\prime$ and $u^\prime$ of $q^\prime$ are non-empty
($\not=\!\eps$) then they are incompatible
and the path that we are following is invalid.
If either $s^\prime$ or $u^\prime$ is empty ($=\!\eps$)
then the current path is valid (at least up to this point)
(Line~\ref{pc:AutoInt:L108}).
Only in this case
and only if the delay between $s^\prime$ and $u^\prime$ does not exceed $\delta_{\rm max2}$,
we construct a transition $e$ in $A$ corresponding to $e_1$ in $A_1$
(Line~\ref{pc:AutoInt:L108},~\ref{pc:AutoInt:L110}).
If its target $q^\prime\!=\!n(e)$ does not exist yet,
it is created and pushed onto the stack
(Lines~\ref{pc:AutoInt:L109}, \ref{pc:AutoInt:L301}--\ref{pc:AutoInt:L310}).
The infinite unrolling of cycles is prevented by $\delta_{\rm max2}$.		\\

{\bf Verification:}~
To see whether the construction was successful
and whether $A\tapnum{n}\!=\!\AInt{j,k}(A_1\tapnum{n})$,
we have to check for the above defined conditions.
Since all states of $A\tapnum{n}$ are reachable,
it is sufficient to verify their delay and coreachability
(Line~\ref{pc:AutoInt:L112})~:
$ \not\!\exists q : |\delta(q)| > \delta_{\rm max}  \logAnd  q {\sf ~coreachable} $.


\RSubSec{Examples}

We illustrate the algorithm through the following three examples
that stand each for a different class of WMTAs.


{\bf Example 1:}~
The relation of the WMTA, $A_1\tapnum{3}$,
of the first example is the infinite set of string tuples
$\aSet{\aTuple{ a b^k , x y^k z , a^k b } | k \in \srSetN}$
~(Figure~\ref{fig:AInt:A11}).
Only one of those tuples, namely $\aTuple{ a b , x y z , a b }$,
is in the relation of the auto-intersection,
$A\tapnum{3}=\AInt{1,3}(A_1\tapnum{3})$,
because all other tuples contain different strings on tape $1$ and $3$.
In the construction, an infinite unrolling of the cycle is prevented
by the incompatibility of the leftover substrings in $\xi[3]$ and $\xi[4]$ respectively.
The construction is successful.

The example is characterized by:
\begin{eqnarray}
  \delta_{\rm max} \;\; = \;\;  \delta_{\rm max2}  & = &  1	\\
  \relat(A_1\tapnum{3}) & = &
	\aSet{\aTuple{ a b^k , x y^k z , a^k b } \;|\; k \in \srSetN}	\\
  \AInt{1,3}(\relat(A_1\tapnum{3})) \; = \; \relat(A\tapnum{3}) & = &
	\aSet{\aTuple{ a b^1 , x y^1 z , a^1 b }}				\\
  \not\!\exists q\!\in\!Q :
	|\delta(\xi[q])|>\delta_{\rm max}
	& \Rightarrow &	{\it successful}
		\spc{2ex}\Rightarrow\spc{2ex}	{\it rational}\;\AInt{1,3}(\;)
\end{eqnarray}

\begin{figure}[ht]  
\begin{center}
\begin{minipage}{40mm}
\includegraphics[scale=0.50,angle=0]{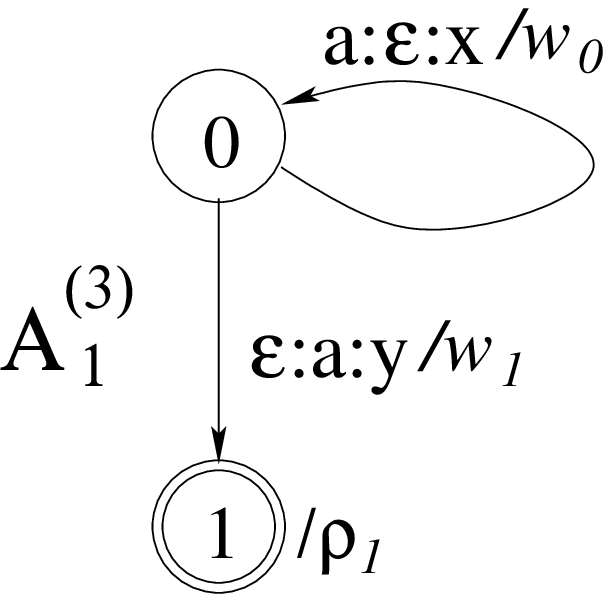}
\end{minipage}
\begin{minipage}{115mm}
\includegraphics[scale=0.50,angle=0]{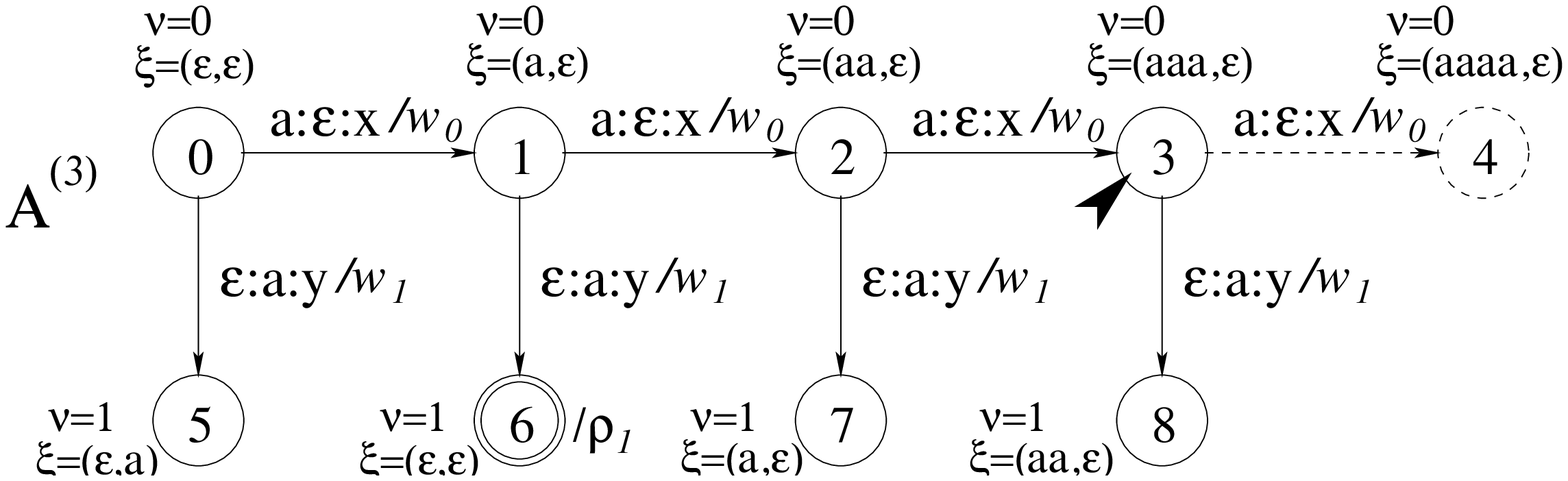}
\end{minipage}

\begin{minipage}{150mm}
\caption{A WMTA $A_1\tapnum{3}$ and its successfully constructed auto-intersection
	$A\tapnum{3}=\AInt{1,2}(A_1\tapnum{3})$.
	(Dashed parts are not constructed.
	States $q$ marked with \blackarrow have $|\delta(\xi[q])|>\delta_{\rm max}$.)
	\label{fig:AInt:A12}}
\end{minipage}
\end{center}
\end{figure}

\begin{figure}[ht]  
\begin{center}
\begin{minipage}{35mm}
\includegraphics[scale=0.50,angle=0]{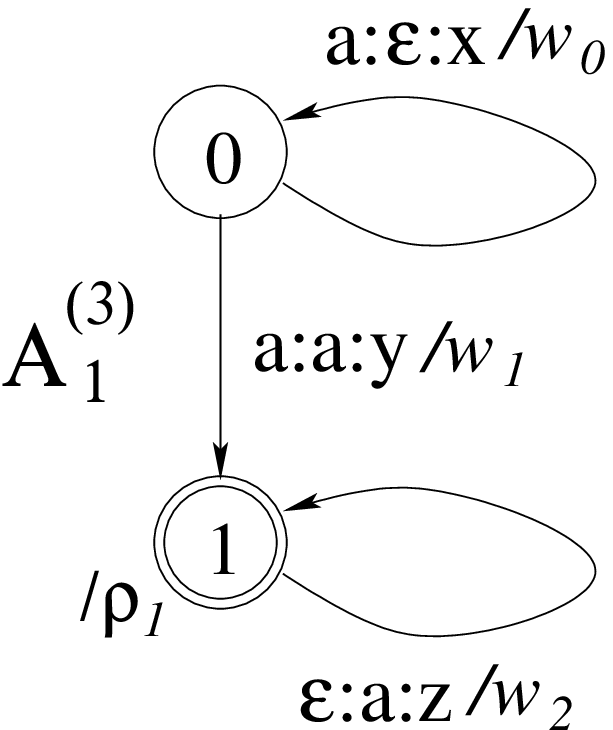}

\vspace{15mm}
\end{minipage}
\begin{minipage}{115mm}
\includegraphics[scale=0.50,angle=0]{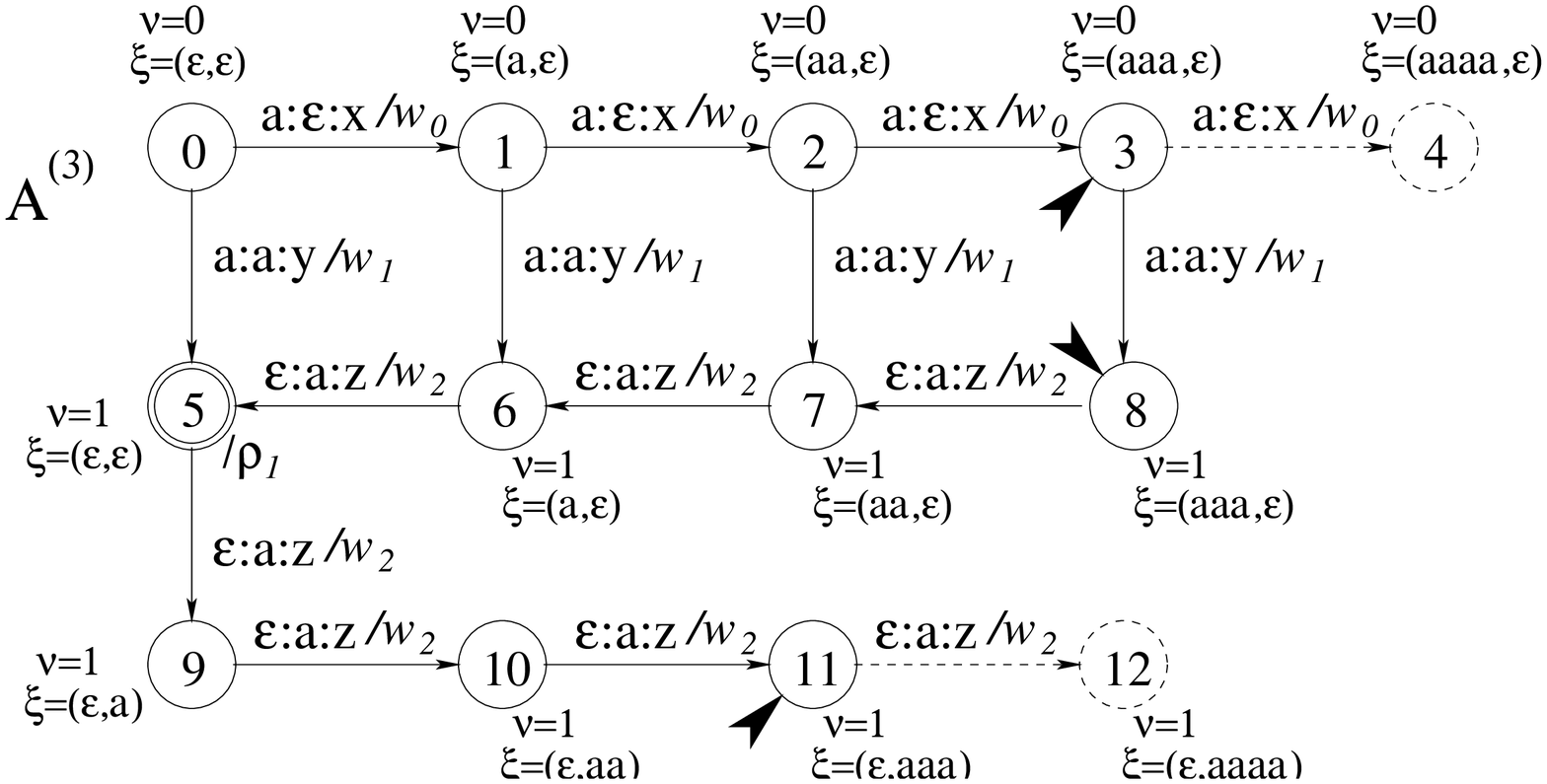}
\end{minipage}

\begin{minipage}{150mm}
\caption{A WMTA $A_1\tapnum{3}$
	and its partially constructed auto-intersection
	$A\tapnum{3}\subset\AInt{1,2}(A_1\tapnum{3})$.
	(Dashed parts are not constructed.
	States $q$ marked with \blackarrow have $|\delta(\xi[q])|>\delta_{\rm max}$.)
	\label{fig:AInt:A13}}
\end{minipage}
\end{center}
\end{figure}

{\bf Example 2:}~
In the second example
(Figure~\ref{fig:AInt:A12}),
the relation of $A_1\tapnum{3}$ is the infinite set of string tuples
$\aSet{\aTuple{ a^k , a , x^k y } \;|\; k \in \srSetN}$.
Only one of those tuples, namely $\aTuple{ a^1 , a , x^1 y }$,
is in the relation of the auto-intersection
$A\tapnum{3}=\AInt{1,2}(A_1\tapnum{3})$.
In the construction, an infinite unrolling of the cycle is prevented
by the limit of delay $\delta_{\rm max2}$.
Although the result contains states with $\delta(\xi[q])|>\delta_{\rm max}$,
none of them is coreachable
(and would disappear if the result was pruned).
The construction is successful.

The example is characterized by:
\begin{eqnarray}
  \delta_{\rm max} & = &	2	\\
  \delta_{\rm max2} & = &	3	\\
  \relat(A_1\tapnum{3}) & = &
	\aSet{\aTuple{ a^k , a , x^k y } \;|\; k \in \srSetN}	\\
  \AInt{1,2}(\relat(A_1\tapnum{3})) \; = \; \relat(A\tapnum{3}) & = &
	\aSet{\aTuple{ a^1 , a , x^1 y }}				\\
  \not\!\exists q\!\in\!Q :
	|\delta(\xi[q])|>\delta_{\rm max} \;\logAnd\; \FCT{coreachable}{q}
	& \Rightarrow &	{\it successful}
		\spc{2ex}\Rightarrow\spc{2ex}	{\it rational}\;\AInt{1,2}(\;)
\end{eqnarray}

{\bf Example 3:}~
In the third example
(Figure~\ref{fig:AInt:A13}),
the relation of $A_1\tapnum{3}$ is the infinite set of string tuples
$\aSet{\aTuple{ a^k a , a a^h , x^k y z^h } \;|\; k,h \in \srSetN}$.
The auto-intersection, $\AInt{1,2}(A_1\tapnum{3})$,
is not rational and has unbounded delay.
Its complete construction would require an infinite unrolling of the cycles of $A_1\tapnum{3}$
and an infinite number of states in $A\tapnum{3}$
which is prevented by $\delta_{\rm max2}$.
The construction is not successful
because the result contains coreachable states with  $\delta(\xi[q])|>\delta_{\rm max}$.

The example is characterized by:
\begin{eqnarray}
  \delta_{\rm max} & = &	2	\\
  \delta_{\rm max2} & = &	3	\\
  \relat(A_1\tapnum{3}) & = &
	\aSet{\aTuple{ a^k a , a a^h , x^k y z^h } \;|\; k,h \in \srSetN}	\\
  \AInt{1,2}(\relat(A_1\tapnum{3})) & = &
	\aSet{\aTuple{ a^k a , a a^k , x^k y z^k } \;|\; k \in \srSetN}	\\
  \AInt{1,2}(\relat(A_1\tapnum{3})) \; \supset \; \relat(A\tapnum{3}) & = &
	\aSet{\aTuple{ a^k a , a a^k , x^k y z^k } \;|\; k \in \rangeL 0, 3 \rangeR }	\\
  \exists q\!\in\!Q :
	|\delta(\xi[q])|>\delta_{\rm max} \;\logAnd\; \FCT{coreachable}{q}
	& \Rightarrow &	{\it not~successful}
\end{eqnarray}

\RSec{Single-Tape Intersection
	\label{sec:Alg:SgInt}}

We propose an algorithm that performs single-tape intersection of two WMTAs,
$A_1\tapnum{n}$ and $A_2\tapnum{m}$, in one step.
Instead of first building the cross-product, $A_1\tapnum{n} \CrPr A_2\tapnum{m}$,
and then deleting most of its paths by auto-intersection, $\AInt{j,n+k}(\;)$,
according to the above procedure (Eq.~\ref{eq:Op:MltInt:proc}),
the algorithm constructs only the useful part of the cross-product.
It is very similar to classical composition of two transducers,
and incorporates the idea of using an $\eps$-filter
in the composition of transducers containing $\eps$-transitions
\cite[Figure~10]{mohri+al:1998}
that will be explained below.
Instead of explicitly using an $\eps$-filter,
we simulate its behaviour in the algorithm.
We will refer to the algorithm as \FUNCT{IntersectCrossEps}{$A_1, A_2, j, k$}:
\begin{eqnarray}
  \mbox{\FUNCT{IntersectCrossEps}{$A_1, A_2, j, k$}}  & = &
	\AInt{j,n+k}(\; A_1\tapnum{n} \CrPr A_2\tapnum{m} \;)  \\
  A_1\tapnum{n} \TInt{j,k} A_2\tapnum{m}  & = &
  	\CProj{n+k} \left(\; \mbox{\FUNCT{IntersectCross}{$A_1, A_2, j, k$}} \;\right)
\end{eqnarray}

\noindent
The complementary projection, $\CProj{n+k}(\;)$,
could be easily integrated into the algorithm
in order to avoid an additional pass.
We keep it apart because \FUNCT{IntersectCrossEps}{~}
serves also as a building block of another algorithm
where this projection must be postponed.


\RSubSec{Mohri's $\eps$-Filter}

\begin{figure}[ht]  
\begin{center}
\begin{minipage}{80mm}
$A_\eps$ \\
\vspace{-5ex}

\includegraphics[scale=0.50,angle=0]{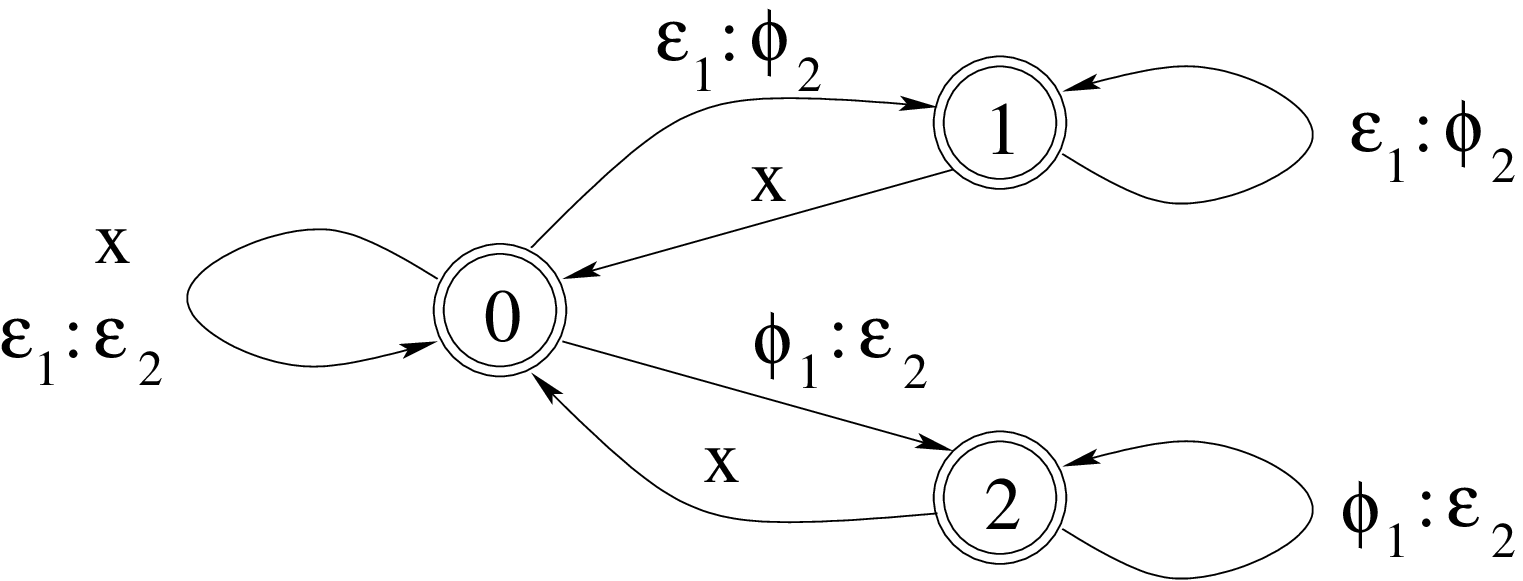}
\end{minipage}

\vspace{4ex}

$A_1$ \includegraphics[scale=0.50,angle=0]{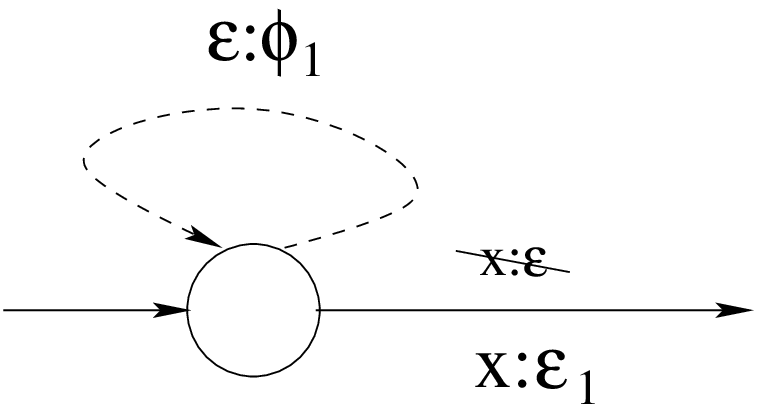}
\spc{15mm}
$A_2$ \includegraphics[scale=0.50,angle=0]{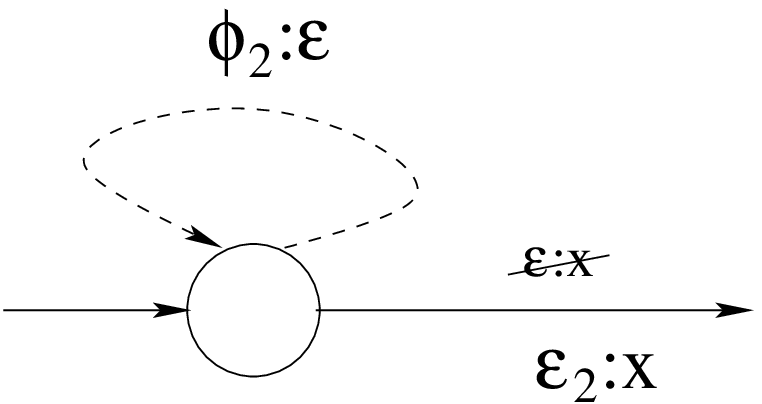}

\begin{minipage}{150mm}
\caption{Mohri's $\eps$-filter $A_\eps$ and two transducers, $A_1$ and $A_2$,
	pre-processed for filtered composition.
	~$x=\neg\aSet{\phi_1, \phi_2, \eps_1, \eps_2}$.
	(For didactic reasons we are using slightly different labels
	than Mohri {\it et al}\/).
	\label{fig:epsFilter}}
\end{minipage}
\end{center}
\end{figure}

To compose two transducers, $A_1\tapnum{2}$ and $A_2\tapnum{2}$ ,
containing $\eps$-transitions,
\namecite[Figure~10]{mohri+al:1998}
are using an $\eps$-filter transducer.
In their approach, $A_1\tapnum{2}$ and $A_2\tapnum{2}$ are pre-processed
(Figure~\ref{fig:epsFilter})~:
each $\eps$ on tape 2 of $A_1\tapnum{2}$ is replaced by an $\eps_1$
and each $\eps$ on tape 1 of $A_2\tapnum{2}$ by an $\eps_2$.
In addition, a looping transition labeled with $\eps\!:\!\phi_1$
is added to each state of $A_1\tapnum{2}$,
and a loop labeled with $\phi_2\!:\!\eps$ to each state of $A_2\tapnum{2}$.
The pre-processed transducers are then composed with the filter $A_\eps\tapnum{2}$
in between: $A_1 \fsCompose A_\eps \fsCompose A_2$.

The filter controls how $\eps$-transitions are composed along each pair of paths
in $A_1$ and $A_2$ respectively.
As long as there are equal symbols ($\eps$ or not) on the two paths,
they are composed with each other and we stay in state 0 of $A_\eps$.
If we encounter a sequence of $\eps$ in $A_1$ but not in $A_2$,
we move forward in $A_1$, stay in the same state in $A_2$, and in state 1 of $A_\eps$.
If we encounter a sequence of $\eps$ in $A_2$ but not in $A_1$,
we move forward in $A_2$, stay in the same state in $A_1$, and in state 2 of $A_\eps$.


\RSubSec{Conditions}

Our algorithm requires the semirings of the two WMTAs
to be equal ($\srK_1\!=\!\srK_2$) and commutative.
All transitions must be labeled with $n$-tuples of strings not exceeding length 1
on the intersected tapes $j$ of $A_1$ and $k$ of $A_2$
which means no loss of generality:~
$
\forall e_1\!\in\!E_1 : |\lab_j(e_1)| \leq 1 ~;~
\forall e_2\!\in\!E_2 : |\lab_k(e_2)| \leq 1
$


\RSubSec{Algorithm}

We start with a WMTA $A$ whose alphabet is the union of the alphabets of $A_1$ and $A_2$,
whose semiring equals those of $A_1$ and $A_2$,
and that is otherwise empty
(Line~\ref{pc:IntCrEps:L1}).

\input{IntCrossEps.pc}

First, we create the initial state $i$ of $A$ from the initial states of $A_1$, $A_2$, and $A_\eps$,
and push $i$ onto the stack
(Lines~\ref{pc:IntCrEps:L3}, \ref{pc:IntCrEps:L26}--\ref{pc:IntCrEps:L32}).
While the stack is not empty,
we take states $q$ from it and access the states $q_1$, $q_2$, and $q_\eps$
that are assigned to $q$ through $\vartheta[q]$
(Lines~\ref{pc:IntCrEps:L5},~\ref{pc:IntCrEps:L6}).

We intersect each outgoing transition $e_1$ of $q_1$
with each outgoing transition $e_2$ of $q_2$
(Lines~\ref{pc:IntCrEps:L7a},~\ref{pc:IntCrEps:L7b}).
This succeeds only if the $j$-th label component of $e_1$
equals the $k$-th label component of $e_2$,
where $j$ and $k$ are the two intersected tapes of $A_1$ and $A_2$ respectively,
and if the corresponding transition in $A_\eps$ has target 0
(Line~\ref{pc:IntCrEps:L8}).
Only if it succeeds, we create a transition in $A$
(Line~\ref{pc:IntCrEps:L12})
whose label results from pairing $\lab(e_1)$ with $\lab(e_2)$ and
whose target $q^\prime$ corresponds with the triple of targets $(n(e_1),n(e_2),0)$.
If $q^\prime$ does not exist yet,
it is created and pushed onto the stack
(Lines~\ref{pc:IntCrEps:L26}--\ref{pc:IntCrEps:L32}).

Subsequently, we handle all $\eps$-transitions in $A_1$
(Lines~\ref{pc:IntCrEps:L13}--\ref{pc:IntCrEps:L18})
and in $A_2$
(Lines~\ref{pc:IntCrEps:L19}--\ref{pc:IntCrEps:L24}).
If we encounter an $\eps$ in $A_1$ and are in state 0 or 1 of $A_\eps$,
we have to move forward in $A_1$, stay in the same state in $A_2$,
and go to state 1 in $A_\eps$.
Therefore we create a transition in $A$ whose target corresponds to the triple
$(n(e_1), q_2, 1)$~
(Lines~\ref{pc:IntCrEps:L13}--\ref{pc:IntCrEps:L18}).
The algorithm works similarly if and $\eps$ is encountered in $A_2$
(Lines~\ref{pc:IntCrEps:L19}--\ref{pc:IntCrEps:L24}).

To adapt this algorithm to non-weighted MTAs,
one has to remove the weights from the
Lines~\ref{pc:IntCrEps:L12},~\ref{pc:IntCrEps:L18}, and~\ref{pc:IntCrEps:L24},
and replace Line~\ref{pc:IntCrEps:L29} with:
$Final(q)$ \SETTO $Final(q_1) \logAnd Final(q_2)$.


\RSec{Multi-Tape Intersection
	\label{sec:Alg:MultInt}}

We propose two alternative algorithms for the multi-tape intersection of two WMTAs,
$A_1\tapnum{n}$ and $A_2\tapnum{m}$.

\RSubSec{Conditions}

Both algorithms work under the conditions of their underlying basic operations:
The semirings of the two WMTAs must be equal ($\srK_1\!=\!\srK_2$) and commutative.
The second (more efficient algorithm) requires
all transitions to be labeled with $n$-tuples of strings not exceeding length 1
on (at least) one pair of intersected tapes
$j_i$ of $A_1\tapnum{n}$ and $k_i$ of $A_2\tapnum{m}$
which means no loss of generality:~
$
\exists i\!\in\!\rangeL 1 , r \rangeR : 
  (\, \forall e_1\!\in\!E_1 : |\lab_{j_i}(e_1)| \leq 1 \,)  \logAnd
  (\, \forall e_2\!\in\!E_2 : |\lab_{k_i}(e_2)| \leq 1 \,)
$

\RSubSec{Algorithms}

Our first algorithm, that we will refer to as
\FUNCT{Intersect1}{$A_1\tapnum{n}, A_2\tapnum{m}, j_1 \dots j_r, k_1 \dots k_r$},
follows the exact procedure of multi-tape intersection
(Eq.~\ref{eq:Op:MltInt:proc}),
using the algorithms for cross product, auto-intersection,
and complementary projection.

\input{MultInt1.pc}

The second (more efficient) algorithm, that we will call
\FUNCT{Intersect2}{$A_1\tapnum{n}, A_2\tapnum{m}, j_1 \dots j_r, k_1 \dots k_r$},
uses first the above single-tape intersection algorithm
to perform cross product and one auto-intersection in one single step
(for intersecting tape $j_1$ with $k_1$),
and then the auto-intersection algorithm
(for intersecting all remaining tapes $j_i$ with $k_i$, for $i>1$).

\input{MultInt2.pc}

This second algorithm has been used to compile successfully the example
of transducer intersection in Section~\ref{sec:Exm:2tapeIntersect}.


\renewcommand{\RSec}[1]{\section{#1}}
\renewcommand{\RSubSec}[1]{\subsection{#1}}
\renewcommand{\RSubSubSec}[1]{\subsubsection{#1}}

\RSec{Applications}

Many applications of WMTAs and WMTA operations are possible,
such as the morphological analysis of Semitic languages
or the extraction of words from a bi-lingual dictionary
that have equal meaning and similar form in the two languages (cognates).

We include only one example in this report,
namely the preservation of intermediate results in transduction cascades,
which actually stands for a large class of applications.


\renewcommand{\RSec}[1]{\subsection{#1}}
\renewcommand{\RSubSec}[1]{\subsubsection{#1}}
\renewcommand{\RSubSubSec}[1]{\subsubsection{SUB -- #1}}


\subsection{Preserving Intermediate Transduction Results
	\label{sec:Apl:TransCasc}}

Transduction cascades have been extensively used in language and speech processing
\cite[among many others]{ait+chanod:1997,pereira+al:1997,kempe:2000,kumar+byrne:2003,kempe+al:2003}.

\begin{figure}[ht]  
\begin{center}
\includegraphics[scale=0.60,angle=0]{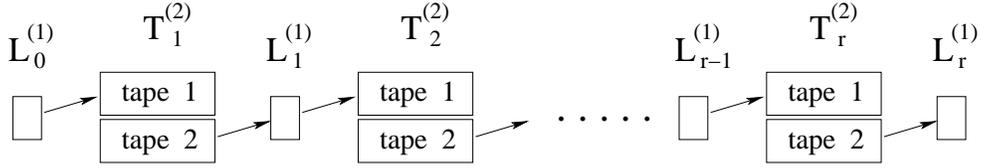}

\caption{Weighted transduction cascade (classical)
		\label{fig:TransCascade}}
\end{center}
\end{figure}

In a (classical) weighted transduction cascade, $T_1\tapnum{2} \dots\; T_r\tapnum{2}$,
a set of weighted input strings, encoded as a weighted acceptor, $L_0\tapnum{1}$,
is composed with the first transducer, $T_1\tapnum{2}$, on its input tape
(Figure~\ref{fig:TransCascade}).
The output projection of this composition is the first intermediate result,
$L_1\tapnum{1}$, of the cascade.
It is further composed with the second transducer, $T_2\tapnum{2}$,
which leads to the second intermediate result, $L_2\tapnum{1}$, etc.
The output projection of the last transducer is the final result, $L_r\tapnum{1}$~:
\begin{equation}
  L_i\tapnum{1} \;\; = \;\; \Proj{2}(\; L_{i-1}\tapnum{1} \fsCompose T_i\tapnum{2} \;)
	\spc{10ex} {\rm for} \spc{2ex}  i \in \rangeL 1,r \rangeR
\end{equation}

\noindent
At any point in the cascade, previous results cannot be accessed.
This holds also if the cascade is composed into a single transducer, $T\tapnum{2}$.
None of the ``incorporated'' sub-relations in $T\tapnum{2}$
can refer to a sub-relation other than its immediate predecessor:
\begin{equation}
  T\tapnum{2} \;\; = \;\; T_1\tapnum{2} \fsCompose \dots \fsCompose T_r\tapnum{2}
\end{equation}

\begin{figure}[ht]	
\begin{center}
\includegraphics[scale=0.60,angle=0]{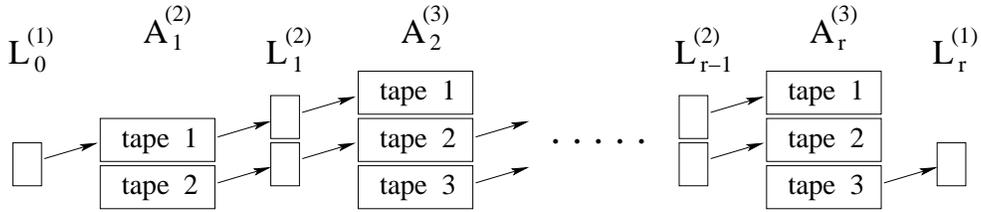}

\caption{Weighted transduction cascade using multi-tape intersection (Example~1)
		\label{fig:WmtaTransCascade}}
\end{center}
\end{figure}

In a weighted transduction cascade, $A_1\tapnum{n_1} \dots\; A_r\tapnum{n_r}$,
that uses WMTAs and multi-tape intersection,
intermediate results can be preserved and used by all subsequent transductions.
Suppose, we want to use the two previous results at each point in the cascade
(except in the first transduction)
which requires all intermediate results, $L_i\tapnum{2}$, to have two tapes
(Figure~\ref{fig:WmtaTransCascade})~:
The projection of the output-tape of the last WMTA is the final result,
$L_r\tapnum{1}$~:
\begin{eqnarray}
  L_1\tapnum{2}  & = &
	L_0\tapnum{1} \TInt{1,1} A_1\tapnum{2}	\\
  L_i\tapnum{2}  & = &
	\Proj{2,3}(\; L_{i-1}\tapnum{2} \TTInt{1,1}{2,2} A_i\tapnum{3} \;)
	\spc{10ex} {\rm for} \spc{2ex}  i \in \rangeL 2, r\!\!-\!\!1 \rangeR	\\
  L_r\tapnum{1}  & = &
	\Proj{3}(\; L_{r-1}\tapnum{2} \TTInt{1,1}{2,2} A_r\tapnum{3} \;)
\end{eqnarray}

This augmented descriptive power is also available
if the whole cascade is intersected into a single WMTA, $A\tapnum{2}$,
although $A\tapnum{2}$ has only two tapes in our example.
This can be achieved by intersecting iteratively the first $i$ WMTAs
until $i$ reaches $r$~:
\begin{equation}
  A_{1\dots i}\tapnum{3}	=
	\Proj{1,n-1,n}(\; A_{1\dots i-1}\tapnum{m} \TTInt{n\!\!-\!\!1,1}{n,2} A_i\tapnum{3} \;)
	\spc{10ex} {\rm for} \spc{2ex}  i\in\rangeL 2,r \rangeR\,,\; m\in\aSet{2,3}
\end{equation}

\noindent
Each $A_{1\dots i}\tapnum{3}$ contains all WMTAs from $A_1\tapnum{2}$ to $A_i\tapnum{3}$.
The final result $A\tapnum{2}$ is built from $A_{1\dots r}\tapnum{3}$~:
\begin{equation}
  A\tapnum{2}	= \Proj{1,n}(\; A_{1\dots r} \;)
\end{equation}

Each (except the first) of the ``incorporated'' multi-tape sub-relations in $A\tapnum{2}$
will still refer to its two predecessors.


\pagebreak

\begin{figure}[ht]	
\begin{center}
\includegraphics[scale=0.60,angle=0]{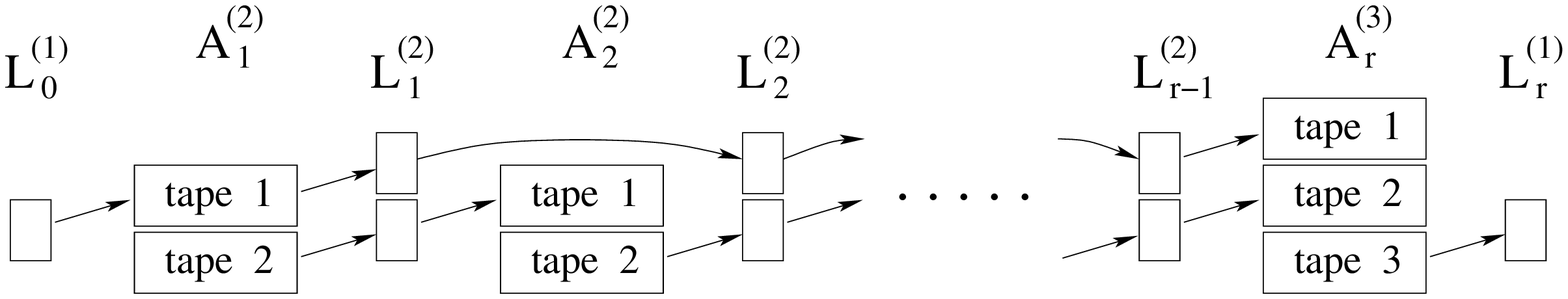}

\caption{Weighted transduction cascade using WMTAs (Example~2)
		\label{fig:WmtaTransCascade2}}
\end{center}
\end{figure}

In our second example of a WMTA cascade, $A_1\tapnum{n_1} \dots\; A_r\tapnum{n_r}$,
each WMTA uses the output of its immediate predecessor,
as in a classical cascade
(Figure~\ref{fig:WmtaTransCascade2}).
In addition, the last WMTA uses the output of the first one:
\begin{eqnarray}
  L_1\tapnum{2}  & = &
	L_0\tapnum{1} \TInt{1,1} A_1\tapnum{2}	\\
  L_i\tapnum{2}  & = &
	\Proj{1,3}(\; L_{i-1}\tapnum{2} \TInt{2,1} A_i\tapnum{2} \;)
	\spc{10ex} {\rm for} \spc{2ex}  i \in \rangeL 2, r\!\!-\!\!1 \rangeR	\\
  L_r\tapnum{1}  & = &
	\Proj{3}(\; L_{r-1}\tapnum{2} \TTInt{1,1}{2,2} A_r\tapnum{3} \;)
\end{eqnarray}

As in the previous example,
the cascade can be intersected into a single WMTA, $A\tapnum{2}$,
that exceeds the power of a classical transducer cascade,
although it has only two tapes:
\begin{eqnarray}
  A_{1\dots i}\tapnum{2}	& = &
	\Proj{1,3}(\; A_{1\dots i-1}\tapnum{2} \TInt{2,1} A_i\tapnum{2} \;)
	\spc{10ex} {\rm for} \spc{2ex}  i\in\rangeL 2, r\!\!-\!\!1 \rangeR	\\
  A_{1\dots r}\tapnum{3}	& = &
	\Proj{1,3}(\; A_{1\dots r-1}\tapnum{2} \TTInt{1,1}{2,2} A_r\tapnum{3} \;)	\\
  A\tapnum{2}			& = &
	\Proj{1,3}(\; A_{1\dots r}\tapnum{3} \;)
\end{eqnarray}


\newpage

\section*{Acknowledgements}

We wish to thank several anonymous reviewers.




\begin{thebibliography}{}

\bibitem[\protect\citename{A\"{\i}t-Mokhtar and Chanod}1997]{ait+chanod:1997}
A\"{\i}t-Mokhtar, Salah and Jean-Pierre Chanod.
\newblock 1997.
\newblock Incremental finite-state parsing.
\newblock In {\em Proc. 5th Int. Conf. ANLP}, pages 72--79, Washington, DC,
  USA.

\bibitem[\protect\citename{Beesley and Karttunen}2003]{beesley+karttunen:2003}
Beesley, Kenneth~R. and Lauri Karttunen.
\newblock 2003.
\newblock {\em Finite State Morphology}.
\newblock CSLI Publications, Palo Alto, CA.

\bibitem[\protect\citename{Birkhoff and Bartee}1970]{birkhoff+bartee:1970}
Birkhoff, Garrett and Thomas~C. Bartee.
\newblock 1970.
\newblock {\em Modern Applied Algebra}.
\newblock McGraw-Hill, New York, NY, USA.

\bibitem[\protect\citename{Eilenberg}1974]{eilenberg:1974}
Eilenberg, Samuel.
\newblock 1974.
\newblock {\em Automata, Languages, and Machines}, volume~A.
\newblock Academic Press, San Diego, CA, USA.

\bibitem[\protect\citename{Elgot and Mezei}1965]{elgot+mezei:1965}
Elgot, Calvin~C. and Jorge~E. Mezei.
\newblock 1965.
\newblock On relations defined by generalized finite automata.
\newblock {\em IBM Journal of Research and Development}, 9(1):47--68.

\bibitem[\protect\citename{Frougny and
  Sakaro\-vitch}1993]{frougny+sakarovitch:1993}
Frougny, Christiane and Jacques Sakarovitch.
\newblock 1993.
\newblock Synchronized rational relations of finite and infinite words.
\newblock {\em Theoretical Computer Science}, 108(1):45--82.

\bibitem[\protect\citename{Ganchev, Mihov, and Schulz}2003]{ganchev+al:2003}
Ganchev, Hristo, Stoyan Mihov, and Klaus~U. Schulz.
\newblock 2003.
\newblock One-letter automata: {How} to reduce $k$ tapes to one.
\newblock CIS-Bericht 03-133, Centrum f\"ur Informations- und
  Sprachverarbeitung, Universit\"at M\"unchen.

\bibitem[\protect\citename{Harju and Karhum\"aki}1991]{harju+karhumaki:1991}
Harju, Tero and Juhani Karhum\"aki.
\newblock 1991.
\newblock The equivalence problem of multitape finite automata.
\newblock {\em Theoretical Computer Science}, 78(2):347--355.

\bibitem[\protect\citename{Kaplan and Kay}1981]{kaplan+kay:1981}
Kaplan, Ronald~M. and Martin Kay.
\newblock 1981.
\newblock Phonological rules and finite state transducers.
\newblock In {\em Winter Meeting of the Linguistic Society of America}, New
  York, NY, USA.

\bibitem[\protect\citename{Kaplan and Kay}1994]{kaplan+kay:1994}
Kaplan, Ronald~M. and Martin Kay.
\newblock 1994.
\newblock Regular models of phonological rule systems.
\newblock {\em Computational Linguistics}, 20(3):331--378.

\bibitem[\protect\citename{Karttunen \bgroup et al.\egroup
  }1997]{karttunen+al:1997}
Karttunen, Lauri, Jean-Pierre Chanod, Greg Grefenstette, and Anne Schiller.
\newblock 1997.
\newblock Regular expressions for language engineering.
\newblock {\em Journal of Natural Language Engineering}, 2(4):307--330.

\bibitem[\protect\citename{Kay}1987]{kay:1987}
Kay, Martin.
\newblock 1987.
\newblock Nonconcatenative finite-state morphology.
\newblock In {\em Proc. 3rd Int. Conf. EACL}, pages 2--10, Copenhagen, Denmark.

\bibitem[\protect\citename{Kempe}2000]{kempe:2000}
Kempe, Andr\'e.
\newblock 2000.
\newblock Reduction of intermediate alphabets in finite-state transducer
  cascades.
\newblock In {\em Proc. 7th Conf. TALN}, pages 207--215, Lausanne, Switzerland,
  October. ATALA.

\bibitem[\protect\citename{Kempe \bgroup et al.\egroup }2003]{kempe+al:2003}
Kempe, Andr\'e, Christof Baeijs, Tam\'as Ga\'al, Franck Guingne, and Florent
  Nicart.
\newblock 2003.
\newblock {WFSC} -- {A} new weighted finite state compiler.
\newblock In O.~H. Ibarra and Z.~Dang, editors, {\em Proc. 8th Int. Conf.
  CIAA}, volume 2759 of {\em Lecture Notes in Computer Science}, pages
  108--119, Santa Barbara, CA, USA. Springer Verlag, Berlin, Germany.

\bibitem[\protect\citename{Kiraz}1997]{kiraz:1997}
Kiraz, George~Anton.
\newblock 1997.
\newblock Linearization of nonlinear lexical representations.
\newblock In John Coleman, editor, {\em Proc. 3rd Meeting, ACL Special Interest
  Group in Computational Phonology}, Madrid, Spain.

\bibitem[\protect\citename{Kiraz and
  Grimley-Evans}1998]{kiraz+grimley-evans:1998}
Kiraz, George~Anton and Edmund Grimley-Evans.
\newblock 1998.
\newblock Multi-tape automata for speech and language systems: {A} {Prolog}
  implementation.
\newblock In D.~Woods and S.~Yu, editors, {\em Automata Implementation}, volume
  1436 of {\em Lecture Notes in Computer Science}. Springer Verlag, Berlin,
  Germany, pages 87--103.

\bibitem[\protect\citename{Koskenniemi, Tapanainen, and
  Voutilainen}1992]{koskenniemi:1992}
Koskenniemi, Kimmo, Pasi Tapanainen, and Atro Voutilainen.
\newblock 1992.
\newblock Compiling and using finite-state syntactic rules.
\newblock In {\em Proc. 16th Int. Conf. COLING}, volume~1, pages 156--162,
  Nantes, France.

\bibitem[\protect\citename{Kuich and Salomaa}1986]{kuich+salomaa:1986}
Kuich, Werner and Arto Salomaa.
\newblock 1986.
\newblock {\em Semirings, Automata, Languages}.
\newblock Number~5 in EATCS Monographs on Theoretical Computer Science.
  Springer Verlag, Berlin, Germany.

\bibitem[\protect\citename{Kumar and Byrne}2003]{kumar+byrne:2003}
Kumar, Shankar and William Byrne.
\newblock 2003.
\newblock A weighted finite state transducer implementation of the alignment
  template model for statistical machine translation.
\newblock In {\em Proc. Int. Conf. HLT-NAACL}, pages 63--70, Edmonton, Canada.

\bibitem[\protect\citename{Mohri}1997]{mohri:1997}
Mohri, Mehryar.
\newblock 1997.
\newblock Finite-state transducers in language and speech processing.
\newblock {\em Computational Linguistics}, 23(2):269--312.

\bibitem[\protect\citename{Mohri}2002]{mohri:2002}
Mohri, Mehryar.
\newblock 2002.
\newblock Generic epsilon-removal and input epsilon-normalization algorithms
  for weighted transducers.
\newblock {\em International Journal of Foundations of Computer Science},
  13(1):129--143.

\bibitem[\protect\citename{Mohri}2003]{mohri:2003}
Mohri, Mehryar.
\newblock 2003.
\newblock Edit-distance of weighted automata.
\newblock In {\em Proc. 7th Int. Conf. CIAA (2002)}, volume 2608 of {\em
  Lecture Notes in Computer Science}, pages 1--23, Tours, France. Springer
  Verlag, Berlin, Germany.

\bibitem[\protect\citename{Mohri, Pereira, and Riley}1998]{mohri+al:1998}
Mohri, Mehryar, Fernando C.~N. Pereira, and Michael Riley.
\newblock 1998.
\newblock A rational design for a weighted finite-state transducer library.
\newblock {\em Lecture Notes in Computer Science}, 1436:144--158.

\bibitem[\protect\citename{Pereira and Riley}1997]{pereira+al:1997}
Pereira, Fernando C.~N. and Michael~D. Riley.
\newblock 1997.
\newblock Speech recognition by composition of weighted finite automata.
\newblock In Emmanuel Roche and Yves Schabes, editors, {\em Finite-State
  Language Processing}. MIT Press, Cambridge, MA, USA, pages 431--453.

\bibitem[\protect\citename{Rabin and Scott}1959]{rabin+scott:1959}
Rabin, Michael~O. and Dana Scott.
\newblock 1959.
\newblock Finite automata and their decision problems.
\newblock {\em IBM Journal of Research and Development}, 3(2):114--125.

\bibitem[\protect\citename{Roche and Schabes}1997]{roche+schabes:1997}
Roche, Emmanuel and Yves Schabes.
\newblock 1997.
\newblock {\em Finite-State Language Processing}.
\newblock MIT Press, Cambridge, MA, USA.

\bibitem[\protect\citename{Sproat}1992]{sproat:1992}
Sproat, Richard.
\newblock 1992.
\newblock {\em Morphology and Computation}.
\newblock MIT Press, Cambridge, MA, USA.

\bibitem[\protect\citename{van Noord}1997]{vannoord:1997}
van Noord, Gertjan.
\newblock 1997.
\newblock {FSA Utilities}: A toolbox to manipulate finite-state automata.
\newblock In D.~Raymond, D.~Woods, and S.~Yu, editors, {\em Automata
  Implementation}, volume 1260 of {\em Lecture Notes in Computer Science}.
  Springer Verlag, Berlin, Germany.

\end{thebibliography}
\end{document}